\documentclass{article}

\usepackage[final]{cpal_2025}

\usepackage{url}

\usepackage[utf8]{inputenc} 
\usepackage[T1]{fontenc}    
\usepackage{hyperref}       
\usepackage{url}            
\usepackage{booktabs}       
\usepackage{amsfonts}       
\usepackage{nicefrac}       
\usepackage{microtype}      
\usepackage{xcolor}         

\usepackage{diagbox}
\usepackage{caption}
\usepackage{subcaption}
\usepackage{wrapfig}

\usepackage{amsmath}
\usepackage{amssymb}
\usepackage{mathtools}
\usepackage{amsthm}

\theoremstyle{plain}
\newtheorem{theorem}{Theorem}[section]
\newtheorem{proposition}[theorem]{Proposition}
\newtheorem{lemma}[theorem]{Lemma}
\newtheorem{corollary}[theorem]{Corollary}
\theoremstyle{definition}
\newtheorem{definition}[theorem]{Definition}
\newtheorem{assumption}[theorem]{Assumption}
\theoremstyle{remark}

\newcommand{\R}{\mathbb{R}}

\newcommand{\blockcomment}[1]{}

\newcommand{\E}{\mathbb{E}}
\newcommand{\pf}[2]{\frac{\partial #1}{\partial #2}}
\newcommand{\pft}[2]{\frac{\partial^2 #1}{\partial #2^2}}

\newcommand{\supp}[1]{\underset{#1}{\mathrm{~sup~}}}

\usepackage{xcolor,colortbl}
\usepackage{forloop}
\newcounter{loopcntr}

\usepackage{listings}
\newcommand{\squishlist}{
   \begin{list}{$\bullet$}
    { \setlength{\itemsep}{0pt}      \setlength{\parsep}{3pt}
      \setlength{\topsep}{3pt}       \setlength{\partopsep}{0pt}
      \setlength{\leftmargin}{1.5em} \setlength{\labelwidth}{1em}
      \setlength{\labelsep}{0.5em} } }

\newcommand{\squishend}{
    \end{list}  }
    
\title{Towards Unraveling and Improving Generalization in World Models}

\author{%
  Qiaoyi Fang\textsuperscript{1}\thanks{An earlier version of this paper was submitted to NeurIPS and received ratings of (7, 6, 6). The reviewers’ comments and the original draft are available at  \href{https://openreview.net/forum?id=3sWghzJvGd&nesting=2&sort=date-desc}{Openreview}. This version contains minor modifications based on that submission.}, ~Weiyu Du\textsuperscript{2}, Hang Wang\textsuperscript{3} ~Junshan Zhang\textsuperscript{3} \\
  \textsuperscript{1}Department of Computer Science, University of California, Davis \\
  \textsuperscript{2}Google\\
  \textsuperscript{3}Department of Electrical and Computer Engineering, University of California, Davis \\
  \texttt{\{qyfang,whang,jazh\}@ucdavis.edu, weiyuduu@gmail.com}
}

\begin{document}

\maketitle

\begin{abstract}
World models have recently emerged as a promising approach to reinforcement learning (RL), achieving state-of-the-art performance across a wide range of visual control tasks. This work aims to obtain a deep understanding of the robustness and generalization capabilities of world models. Thus motivated, we develop a stochastic differential equation formulation  by treating the world model learning as a stochastic dynamical system, and characterize the impact of latent representation errors on robustness and generalization, for both cases with zero-drift representation errors and  with non-zero-drift representation errors. Our  somewhat surprising findings, based on both theoretic and experimental studies,  reveal that for the case with zero drift, modest latent representation errors can in fact function as implicit regularization and hence result in improved robustness.  We further propose a Jacobian regularization scheme to mitigate the compounding error propagation effects of non-zero drift, thereby enhancing training stability and robustness. Our experimental studies corroborate that this regularization approach not only stabilizes training but also accelerates convergence and improves  accuracy of long-horizon prediction. 
\end{abstract}

\section{Introduction}

Model-based reinforcement learning (RL) has emerged as a promising paradigm to improve sample efficiency by enabling agents to exploit a learned model of the physical environment. Recent works on world models  \cite{hafner2019learning,hafner2020dream,hafner2022mastering,hafner2023mastering,kessler2023effectiveness,freeman2019learning,wu2023daydreamer,kim2020active} 
involve RL agents learning a latent dynamics model (LDM) from observations and actions, and then optimizing the policy over this learned model. Unlike conventional approaches, world-model-based RL employs an {\em end-to-end learning} strategy, jointly training the dynamics model, perception, and action policy to achieve a unified goal.  This framework offers significant potential to improve both generalization and robustness to perturbations, making it highly advantageous for real-world scenarios. For example, DreamerV2-V3 have achieved notable progress in mastering diverse tasks involving continuous and discrete actions, image-based inputs, and both 2D and 3D environments \cite{hafner2019learning,hafner2020dream,hafner2022mastering}. Recent empirical studies have also demonstrated the capacity of world models to generalize to unseen noisy states and dynamics in complex environments, such as autonomous driving \cite{hu2023gaia1}. However, it remains unclear when and how world models can generalize well in unseen environments, and the role of robustness in this process.

In this work, we aim to systematically understand the \textit{robustness} and \textit{generalization} capabilities of world models by examining the impact of \textit{latent representation errors} introduced by latent encoders. Specifically, we investigate how these errors can enhance robustness against perturbations, which in turn often improves generalization \cite{lim2021noisy}. Contrary to the expectation that minimizing latent representation errors by optimizing the LDM prior to policy training would lead to better performance, our theoretical and empirical findings reveal that modest latent representation errors during training may actually be beneficial for robustness. In particular, the alternating training strategy for world model learning, which simultaneously refines both the LDM and the action policy, can improve robustness and yield generalization gains. This is because modest latent representation errors enable the world model to better handle perturbations, leading to improved exploration and generalization capabilities.

This phenomenon mirrors the behavior observed with gradient estimation errors in batch training. For instance, as shown  in Table \ref{table:expr-batch}, intermediate batch sizes (e.g., 16 or 32) produce gradient estimation errors that are beneficial for generalization, compared to smaller (e.g., 8) or larger (e.g., 64) batch sizes. The latent representation errors exhibit a similar effect in a controlled range, supporting robustness through implicit regularization. Indeed,  implicit regularization has been associated with increased classification margins \cite{poggio2017theory}, which improves generalization performance \cite{sokolic2017generalization, lim2021noisy}.

\begin{table} 
\centering
\begin{tabular}{c|ccc|ccc}
\toprule
\diagbox[height=0.75cm]{batch size}{perturbation}& $\alpha=10$& $\alpha=20$& $\alpha=30$& $\beta=25$& $\beta=50$& $\beta=75$\\
\hline
8&  691.62&    363.73&     153.67&      624.67&     365.31& 216.52\\
16&  \textbf{830.39}&      429.62&    \textbf{213.78}&      \textbf{842.26}&          \textbf{569.42}&    \textbf{375.61}\\

 32& \textbf{869.39}& \textbf{436.87}& \textbf{312.99}& \textbf{912.12}& \textbf{776.86}&\textbf{655.26}\\
 64& 754.47& \textbf{440.44}& 80.24& 590.41& 255.2&119.62\\
 \bottomrule
\end{tabular}
\caption{Reward values on unseen perturbed states by rotation ($\alpha$) or mask ($\beta\%$) with $\mathcal{N}(0.15, 0.5)$.}
\label{table:expr-batch}
\vspace{-12 pt}
\end{table}

In a nutshell, \textit{latent representation errors}, if properly managed, may actually facilitate world model training by enhancing robustness against perturbations, thereby improving generalization.  This insight aligns with recent advances in deep learning, where noise injection schemes have been studied as a form of implicit regularization to enhance model robustness. For instance,  \cite{camuto2021explicit} analyzes the effects of introducing isotropic Gaussian noise at each layer of neural networks, identifying it as implicit regularization. Similarly, \cite{lim2021noisy} explores adding zero-drift Brownian motion to RNN architectures, demonstrating its regularizing effects in improving stability.

However, we caution that \textit{latent representation errors} in world models  differ from the noise injection schemes  (\cite{lim2021noisy,camuto2021explicit}), in several key aspects: 1) Unlike the artificially injected noise only added in training and removed during inference, these errors are inherent in world models and lead to error propagation during rollouts; 2) The errors in world models may not exhibit well-behaved properties such as isotropic or zero-drift noise and may have non-zero drift and bias; 3) in the iterative training of LDMs and agents, the error originating from the encoder also affects the policy learning and exploration.

To address these challenges, we develop a continuous-time stochastic differential equation (SDE) formulation by modeling LDM as a stochastic dynamical system, aiming to  understand the robustness  and generalization  of world models and  improve it further. This approach provides a formal characterization of latent representation errors as stochastic perturbations, allowing us to quantify their impact on robustness and generalization. Our main contributions can be summarized as follows:\label{contribution}
  
\squishlist
    \item \textit{Latent representation errors as implicit regularization:}
  We  develop a continuous-time SDE formulation  by treating the world model learning as  a stochastic dynamical system. Using stochastic perturbation results, we show that under certain conditions, modest latent representation errors can in fact act as implicit regularization, leading to robustness gain. 
 
    \item \textit{Improving robustness and generalization in non-zero drift cases via Jacobian regularization:} 
    For the case where latent representation errors exhibit non-zero drifts, we show that the additional bias can degrade the implicit regularization effect, leading to learning instability. We theoretically quantify this instability and show that the well-known Jacobian regularization can be employed to address this issue, verified by our experimental results. 

    \item \textit{Reducing error propagation in predictive rollouts:} We explicitly characterize the effect of latent representation errors on predictive rollouts and robustness.By applying Jacobian regularization, we control these errors, leading to reduced error propagation, enhanced prediction performance, and faster convergence, especially in tasks with longer time horizons.

       
\squishend

\paragraph{Notation.}
We use the Einstein summation convention for succinctness, where $a_ib_i$ denotes $\sum_i a_ib_i$. We denote functions in $\mathcal{C}^{k,\alpha}$ as being $k$-times differentiable with $\alpha$-Hölder continuity. The Euclidean norm of a vector is represented by $\|\cdot\|$, and the Frobenius norm of a matrix by $|\cdot|_F$; this notation may occasionally extend to tensors. The notation $x^i$ indicates the $i^{th}$ coordinate of the vector $x$ and $A^{ij}$ the $(i,j)$ entry of the matrix $A$. The composition of the function is denoted by $f \circ g$. For a differentiable function $f: \mathbb{R}^{n} \rightarrow \mathbb{R}^{m}$ , its Jacobian matrix is denoted by $\frac{\partial f}{\partial x} \in \mathbb{R}^{m \times n}$. Its gradient, following conventional definitions, is denoted by $\nabla f$. The constant $C$ may represent different values in different contexts.
\raggedbottom

\section{Related Work}
\textbf{Robustness and Generalization in Deep RL.}
Recent work on deep RL robustness and generalization has studied zero-shot generalization of learned policies to unseen environments \cite{kirk2023survey}, often emphasizing task-level generalization through techniques such as task augmentation in meta-RL \cite{yao2021improving, lee2021improving}. In contrast, our work targets the robustness and generalization of world-model-based RL under observational and dynamic perturbations, emphasizing the role of latent representations. While recent studies on RL robustness \cite{panaganti2022robust, liu2023robustness} introduce new training frameworks aimed at policy safety and robustness, they do not account for the inherent challenges posed by latent representation errors during rollouts. 


\textbf{World model based RL.}
World models have excelled in visual control tasks across various platforms, including Atari \cite{bellemare2013arcade} and Minecraft \cite{duncan2011minecraft}, as detailed in the studies by Hafner et al. \cite{hafner2019learning,hafner2020dream,hafner2022mastering}. These models typically integrate encoders and memory-augmented neural networks, such as RNNs \cite{yu2019review}, to manage the latent dynamics. The use of variational autoencoders (VAE) \cite{doersch2016tutorial,kingma2013auto} to map sensory inputs to a compact latent space was pioneered by Ha et al. \cite{ha2018world}. Furthermore, the Dreamer algorithm \cite{hafner2020dream,hafner2023mastering} employs convolutional neural networks (CNNs) \cite{lecun1989backpropagation} to enhance the processing of both hidden states and image embeddings, yielding models with improved predictive capabilities in dynamic environments. 

\textbf{Continuous-time RNNs.}
The continuous-time assumption is standard for theoretical formulations of RNN models. Li et al. \cite{JMLR:v23:21-0368} study the optimization dynamics of linear RNNs on memory decay.  Chang et al. \cite{chang2019antisymmetricrnn} propose AntisymmetricRNN, which captures long-term dependencies through the control of eigenvalues in its underlying ODE. Chen et al. \cite{chen2020symplectic} propose the symplectic RNN to model Hamiltonians. As continuous-time formulations can be discretized with Euler methods \cite{chang2019antisymmetricrnn, chen2020symplectic} (or with Euler-Maruyama methods if stochastic in \cite{lim2021noisy}) and yield similar insights, this step is often eliminated for brevity. 

\textbf{Implicit regularization by noise injection in RNN.}
Studies on noise injection as a form of implicit regularization have gained traction, with Lim et al. \cite{lim2021noisy} deriving an explicit regularizer under small noise conditions, demonstrating bias towards models with larger margins and more stable dynamics. Camuto et al. \cite{camuto2021explicit} examine Gaussian noise injections at each layer of neural networks. Similarly, Wei et al. \cite{wei2020implicit} provide analytic insights into the dual effects of dropout techniques.



\section{Demystifying World Model: A Stochastic Differential Equation Approach}

As pointed out in \cite{hafner2019learning,hafner2020dream,hafner2022mastering,hafner2023mastering}, critical to the effectiveness of the world model representation is the stochastic design of its latent dynamics model. The model can be outlined by the following key components: an encoder that  compresses high dimensional observations $s_t$ into a low-dimensional latent state $z_t$ (Eq.\ref{enc}), a sequence model that captures temporal dependencies in the environment (Eq.\ref{rnn}), a transition predictor that estimates the next latent state (Eq.\ref{trans}), and a latent decoder that reconstructs observed information from the posterior (Eq.\ref{dec}): 
{\footnotesize
\begin{align}
\text{Latent Encoder: } & z_{t} \sim  q_{\text{enc}}(z_{t} \,\vert \,h_{t}, s_t),\label{enc}\\
\text{Sequence Model: } & h_{t} = f(h_{t-1}, z_{t-1}, {a}_{t-1})\label{rnn},\\
\text{Transition Predictor: } & \tilde{z}_t \sim p(\tilde{z_t} \,|\, h_t),\label{trans}\\
\text{Latent Decoder: } & \tilde{s_t} \sim q_{\text{dec}} (\tilde{s_t}\,|\,h_t, z_t)\label{dec}
\end{align}
}
In this work, we consider a popular class of world models, including Dreamer and PlaNet, where \{$z$, $\tilde{z}$, $\tilde{s}$\} have distributions parameterized by neural networks' outputs, and  are   Gaussian   when the outputs are known. It is worth noting that  \{$z$, $\tilde{z}$, $\tilde{s}$\} may not be Gaussian and are non-Gaussian in general. This is because while $z$ is conditional Gaussian,  its mean and variance are random variables   which are learned by the encoder with $s$ and $h$ being the inputs,  rendering that $z$ is non-Gaussian due to the mixture effect.  For this setting, we have a  continuous-time formulation where the latent dynamics model can be interpreted as stochastic differential equations (SDEs) with  coefficient functions of known inputs. Due to space limitation, we refer to Proposition \ref{prop-lat-gau} in the Appendix for a more detailed treatment. 

Consider a complete, filtered probability space $(\Omega, \,\mathcal{F}, \,\{\mathcal{F}_t\}_{t\in[0, T]}, \,\mathbb{P}\,)$ where independent standard Brownian motions $B^{\text{\,enc}}_t,\, B^{\text{\,pred}}_t, B^{\text{\,seq}}_t, \,B^{\text{\,dec}}_t$ are defined such that ${\mathcal{F}_t}$ is their augmented filtration, and $T \in \mathbb{R}$ as the time length of the task environment. We interpret the stochastic dynamics of LDM with latent representation errors through coupled   SDEs  representing continuous-time analogs of the discrete components: 
{\footnotesize
\begin{align}
\text{Latent Encoder: }  & d\,z_t =  \left( q_{\text{enc}}(h_t, s_t) + \varepsilon\,\sigma(h_t, s_t)\right)\,dt  + \left(\bar{q}_{\text{enc}}(h_t, s_t) + \varepsilon\,\bar{\sigma}(h_t, s_t) \right)d B^\text{\,enc}_t,\label{per-encoder}\\
\text{Sequence Model: } & d\,h_t  = f(h_t, z_t, \pi(h_t, z_t)) \, dt +  \bar{f}(h_t, z_t, \pi(h_t, z_t)) \,d B^{\text{\,seq}}_t \label{unper-h}\\
\text{Transition Predictor: } & d\,\tilde{z}_t = p(h_t)\,dt + \bar{p} (h_t)\,dB^{\text{\,pred}}_t,\label{unper-z}\\
\text{Latent Decoder: } &  d\,\tilde{s}_t = q_{\text{dec}}(h_t, \tilde{z}_t)\,dt + \bar{q}_{\text{dec}}(h_t, z_t)\,dB^{\text{\,dec}}_t \label{unper-est-s},
\end{align}
}
where $\pi(h, \tilde{z})$ is a policy function as a local maximizer of value function and the stochastic process $s_t$ is $\mathcal{F}_t$-adapted. Notice that $\bar{f}$ is often a zero function indicating that Equation (\ref{unper-h}) is an ODE, as the sequence model is generally designed as deterministic.  Generally, the coefficient functions in $dt$ and $dB_t$ terms in SDEs 
are referred to as the \textit{drift} and \textit{diffusion} coefficients. Intuitively,  the diffusion coefficients here represent the stochastic model components. 

For latent representation errors, in Equation (\ref{per-encoder}), $\sigma(\cdot, \cdot)$ and $\bar{\sigma}(\cdot, \cdot)$ denotes the drift and diffusion coefficients of the stochastic \textit{latent representation errors}, respectively, both of which depend on hidden states $h_t$ and task states $s_t$. The parameter $\varepsilon$ serves as a scaling factor of the stochastic error. In Appendix A, we provide a theoretical justification (see \ref{approx-thm-main-appendix}) showing that the latent representation error, in the form of approximation error corresponding to the widely used CNN encoder-decoder, could be made sufficiently small by finding appropriate CNN network configuration. In particular, this result justifies interpreting latent representation error as a stochastic perturbation in the dynamical system defined in Equations (\ref{per-encoder} - \ref{unper-est-s}), as the error magnitude $\varepsilon$ can be made sufficiently small by CNN network configuration.


Next, we impose standard assumptions to guarantee the well-definedness of the solution to SDEs. For further technical details, we refer readers to fundamental works on SDEs in the literature (e.g.,\cite{Steele_2001,Hennequin_Dudley_Kunita_Ledrappier_1984}).  
\begin{assumption}\label{reg-sde}
The drift coefficient functions $q_\text{enc},$ $f,$  $p$ and $q_\text{dec}$ and the diffusion coefficient functions $\bar{q}_\text{enc},  \,\bar{p}$ and $\bar{q}_{\text{dec}}$  are bounded and Borel-measurable over the interval $[0, T]$, and of class $\mathcal{C}^3$ with bounded Lipschitz continuous partial derivatives. The initial values $z_0, h_0, \tilde{z}_0, \tilde{s}_0$ are square-integrable random variables. 

\begin{assumption}
$\sigma$ and $\bar{\sigma}$ are bounded and Borel-measurable and are of class $\mathcal{C}^3$ with bounded Lipschitz continuous partial derivatives over the interval $[0, T]$. \label{noise-reg}
\end{assumption}


\end{assumption}

\subsection{Latent Representation Errors as  Implicit Regularization towards Robustness and Generalization}

In this section, we investigate how latent representation errors influence both robustness and generalization, considering two scenarios: \textit{zero drift} and \textit{non-zero drift}. Our analysis shows that under mild conditions, \textit{zero-drift} errors can act as a natural form of \textit{implicit regularization}, creating wider optimization landscapes that enhance robustness. However, when latent representation errors exhibit \textit{non-zero drift}, they introduce an \textit{unstable bias} that undermines the implicit regularization effect, leading to degraded generalization performance. In such cases, explicit regularization is necessary to stabilize learning and maintain both robustness and generalization capabilities in the world model.


To simplify the notation here, we consider the system equations, specifically Equations (\ref{per-encoder}), (\ref{unper-h}) - (\ref{unper-est-s}), as one stochastic system.  Let $x_t = (z_t, h_t, \tilde{z}_t, \tilde{s}_t)$ and  $B_t = (B^{\text{\,enc}}_t, B^{\text{\,seq}}_t, B^{\text{\,pred}}_t, B^{\text{\,dec}}_t)$:
\begin{align}\label{ultSDE}
d\,x_{t} = \left(g(x_{t}, t) + \varepsilon\, \sigma (x_{t}, t) \right) \, dt + \sum_i \bar{g}_i(x_t, t) + \varepsilon\, \bar{\sigma}_i (x_{t}, t)\, d B^i_{t},
\end{align}
where $g$, and $\bar{g}_i$ are structured accordingly for the respective components, employing the Einstein summation convention for concise representation. For abuse of notation, $\sigma = (\sigma, 0, 0, 0), \bar{\sigma} = (\bar{\sigma}, 0, 0, 0).$  
For a given error magnitude $\varepsilon$, we denote the solution to SDE (\ref{ultSDE}) as $x^{\varepsilon}_t$. Intuitively, $x^{\varepsilon}_t$ is the perturbed trajectory of the latent dynamics model. In particular, when $\varepsilon = 0$, indicating that the absence of latent representation error in the model, the solution is denoted as $x_t^0$.

\subsubsection{The Case with Zero-drift  Representation Errors}

When the drift coefficient $\sigma = 0$, the latent representation errors correspond to a class of well-behaved stochastic processes.  The following result translates the induced perturbation on the stochastic latent dynamics model's loss function $\mathcal{L}$ to a form of explicit regularization. We assume that a (nonconvex) general loss function $\mathcal{L} \in \mathcal{C}^2$ which  depends on $z_t, h_t, \tilde{z}_t, \tilde{s}_t$. Loss functions used in practical implementation, e.g. in DreamerV3, reconstruction loss $J_O$, reward loss $J_R$, consistency loss $J_D$, all satisfy this condition.
\begin{theorem} \textbf{(Explicit Effect Induced by Zero-Drift Representation Error)}\label{imp-reg-thm}
Under Assumptions \ref{reg-sde}  and \ref{noise-reg} and considering a loss function $\mathcal{L} \in \mathcal{C}^2$, the explicit effects of the zero-drift error can be marginalized out as follows: as $\varepsilon \rightarrow 0$,
{\footnotesize
\begin{equation}
    \E\, \mathcal{L}\left(x^\varepsilon_t\right) = \E\,\mathcal{L}(x_t^0) +  \mathcal{R}  + \mathcal{O}(\varepsilon^3),
\end{equation}
}
where the regularization term $\mathcal{R}$ is given by  {\footnotesize
$\mathcal{R} := \, \varepsilon\, \mathcal{P} + \varepsilon^2 \left(\mathcal{Q} +  \frac{1}{2}\,\mathcal{S} \right),$ } with
{\footnotesize
\begin{align}
    \mathcal{P} := & \,\E \,\nabla \mathcal{L}(x_t^0)^\top \Phi_t \sum _k \xi_t^k,\\
    \mathcal{S} :=  &\,\E \sum _{k_1, k_2} (\Phi_t \xi_t^{k_1})^i \nabla^2 \mathcal{L}(x_t^0, t)\,(\Phi_t \xi_t^{k_2})^j,\\
    \mathcal{Q} := & \,\E \, \nabla \mathcal{L} (x_t^0)^\top  \Phi_t \int_0^t \Phi_s^{-1} \, \mathcal{H}^k(x_s^0, s) dB^k_t. 
\end{align}
}
Square matrix $\Phi_t$ is the stochastic fundamental matrix of the corresponding homogeneous equation:
{\footnotesize
\begin{equation*}
    d \Phi_t = \frac{\partial \bar{g}_k}{\partial x}(x_t^0, t)\, \Phi_t\, dB^k_t, \quad \Phi(0) = I,
\end{equation*}
}
and $\xi_t^k$ is the shorthand for $\int_0^t \Phi_s^{-1}\bar{\sigma}_k(x_s^0, s)dB^k_t$. Additionally, $\mathcal{H}^k(x_s^0, s)$ is represented by for $\sum _{k_1, k_2} \frac{\partial^2 \bar{g}_{k}}{\partial x^i \partial x^j}(x^0_s, s)\left( \xi_s^{k_1}\right)^i\left( \xi_s^{k_2}\right)^j$.
\end{theorem}
The proof is relegated to Appendix~\ref{explicit reg} in the Supplementary Materials.

In the special case when the loss $\mathcal{L}$ is convex, then its Hessian, $\nabla^2 \mathcal{L}$, is positive semi-definite, which ensures that the term $\mathcal{S}$ is non-negative.  {\em The presence of this Hessian-dependent term $\mathcal{S}$, under latent representation error, implies a tendency towards wider minima in the loss landscape.} Empirical results from \cite{keskar2017largebatch} indicates that wider minima correlate with improved robustness of implicit regularization during training. This observation also aligns with the theoretical insights in \cite{lim2021noisy}  that the introduction of Brownian motion, which is indeed zero-drift by definition, in training RNN models promotes robustness. We note that in addition, when the error $\bar{\sigma}_t(\cdot)$ is too small, the effect of term $\mathcal{S}$ as implicit regularization would not be as significant as desired. Intuitively, this insight resonates with the empirical results in Table \ref{table:expr-batch} that model's robustness gain is not significant when the error induced by large batch sizes is too small. 

We remark that the exact loss form treated here is simplified compared to  that in the practical implementation of world models, which frequently depends on the probability density functions (PDFs) of $z_t, h_t, \tilde{z}_t, \tilde{s}_t$. In principle, the PDE formulation corresponding to  the PDFs of the perturbed  $x_t^\varepsilon$ can be derived from the Kolmogorov equation of the SDE (\ref{ultSDE}), and the technicality is more involved but can offer more direct insight.  We will study this in future work.

\subsubsection{The Case with Non-Zero-Drift Representation Errors}

In practice, latent representation errors may not always exhibit  \textit{zero drift} as in idealized noise-injection schemes for deep learning (\cite{lim2021noisy}, \cite{camuto2021explicit}). When the drift coefficient $\sigma$ is non-zero or a  function of input data $h_t$ and $s_t$ in general, the explicit regularization terms induced by the latent representation error may lead to  unstable bias in addition to the regularization term $\mathcal{R}$  in Theorem \ref{imp-reg-thm}. With a slight abuse of notation, we denote $\bar{g}_0$ as $g$ from Equation (\ref{ultSDE}) for convenience. 
\begin{corollary} \textbf{(Additional Bias Induced by Non-Zero Drift Representation Error)}\label{unstable-imp-thm}\\
Under Assumptions \ref{reg-sde}  and \ref{noise-reg} and considering a loss function $\mathcal{L} \in \mathcal{C}^2$, the explicit effects of the general form error can be marginalized out as follows as $\varepsilon \rightarrow 0$:
{\footnotesize
\begin{equation}
    \E\, \mathcal{L}\left(x^\varepsilon_t\right) = \E\,\mathcal{L}(x_t^0) +  \mathcal{R} + \tilde{\mathcal{R}} + \mathcal{O}(\varepsilon^3),
\end{equation} 
}
  where the additional bias term $\tilde{\mathcal{R}}$ is given by 
$\tilde{\mathcal{R}} := \, \varepsilon\, \tilde{\mathcal{P}} + \varepsilon^2 \left(\tilde{\mathcal{Q}} +  \tilde{\mathcal{S}} \right)$, with  
{\footnotesize
\begin{align}
    \tilde{\mathcal{P}} := & \,\E \,\nabla \mathcal{L}(x_t^0)^\top \Phi_t \, \tilde{\xi}_t,\\
    \tilde{\mathcal{Q}} := & \,\E \, \nabla \mathcal{L} (x_t^0)^\top  \Phi_t \int_0^t \Phi_s^{-1} \, \mathcal{H}^0(x_s^0, s) \,dt, \\
    \tilde{\mathcal{S}} := & \,\E \sum _{k} (\Phi_t \tilde{\xi}_t)^i \nabla^2 \mathcal{L}(x_t^0, t)\,(\Phi_t \xi_t^{k})^j,
\end{align}
}
and $\tilde{\xi}_t$ being   the shorthand for $\int_0^t \Phi_s^{-1}\sigma_k(x_s^0, s)dt$. 
\end{corollary}

The presence of the new bias term $\tilde{\mathcal{R}}$ implies that regularization effects of latent representation error could be unstable. The presence of $\tilde{\xi}$ in $\tilde{\mathcal{P}}$, $\tilde{\mathcal{Q}}$ and $\tilde{\mathcal{S}}$ induces a bias to the loss function with its magnitude dependent on the error level $\varepsilon$, since $\tilde{\xi}$ is a non-zero term influenced on the drift term $\sigma$. This contrasts with the scenarios described in \cite{lim2021noisy} and \cite{camuto2021explicit}, where the noise injected for implicit regularization follows a   zero-mean Gaussian distribution. To modulate the regularization and bias terms $\mathcal{R}$ and $\tilde{\mathcal{R}}$ respectively, we note that a common factor, the fundamental matrix $\Phi$, can be bounded by 
\begin{equation}\label{fund-ineq}
     \E \sup_{t}\|\Phi_t\|^2_F \leq \sum _k  C \exp \left( C \,\E \sup_{t} \left\|\frac{\partial g_k}{\partial x}(x_t^0, t)\right\|^2_F\right)
\end{equation}
which can be shown by  using the Burkholder-Davis-Gundy Inequality and Gronwall's Lemma. Based on this observation, we next propose a regularizer on input-output Jacobian norm  $\|\frac{\partial g_k}{\partial x}\|_F$ that could modulate the new bias term $\tilde{\mathcal{R}}$   for stabilized implicit regularization. 

\section{Enhancing Predictive Rollouts via Jacobian Regularization}


In this section, we study the effects of latent representation errors on predictive rollouts using latent state transitions, which happen in the inference phase in world models. We then propose to use Jacobian regularization to enhance the quality of rollouts. In particular, we first obtain an upper bound of state trajectory divergence in the rollout due to the representation error. We show that the error effects on task policy's $Q$ function can be controlled through model's input-output Jacobian norm.

In world model learning, the task policy is optimized over the rollouts of dynamics model with the initial latent state $z_0$. Recall that latent representation error is introduced to $z_0$ when latent encoder encodes the initial state $s_0$ from task environment. Intuitively, the latent representation error would propagate under the sequence model and impact the policy learning, which would then affect the generalization capacity through increased exploration. 

Recall that the sequence model and the transition predictor are given as follows:
\begin{equation}
        d\,h_t = f(h_t, \tilde{z}_t, \pi(h_t, \tilde{z}_t))\, dt, \quad
    d\,\tilde{z}_t = p(h_t)dt + \bar{p} (h_t)\,dB_t,\label{inf-per-z}
\end{equation}

with random variables $h_0$, $\tilde{z}_0 + \varepsilon$ as the initial values, respectively. In particular, $\varepsilon$ is a random variable of proper dimension, representing the error from encoder introduced at the initial step.  We impose the standard assumption on the error to ensure the well-definedness of the SDEs. 

Under Assumption \ref{reg-sde}, there exists a unique solution to the SDEs (for Equations \ref{inf-per-z} with square-integrable $\varepsilon$), denoted as $(h_t^\varepsilon, z_t^\varepsilon)$. In the case of no error introduced, i.e., $\varepsilon =0$, we denote the solution of the SDEs as $(h_t^0, z_t^0)$  understood as the rollout under the absence of latent representation error.  To understand how to modulate impacts of the error in rollouts, our following result gives an upper bound on the expected divergence between the perturbed rollout trajectory $(h_t^\varepsilon, z_t^\varepsilon)$ and the original $(h_t^0, z_t^0)$ over the interval $[0, T]$.

\begin{theorem}\label{acc-thm}
\textbf{ (Bounding trajectory divergence) }
For a square-integrable random variable $\varepsilon$, {\footnotesize let $\delta := \E \,\|\varepsilon\|$ and $d_{\varepsilon} := \mathbb{E} \sup _{t \in [0, T]} \left\|h_{t}^{\varepsilon}-h_{t}^{0}\right\|^2 + \left\|\tilde{z}_{t}^{\varepsilon}-\tilde{z}_{t}^{0}\right\|^2.$} As $\delta \rightarrow 0$, 
\begin{align*}
d_{\varepsilon} \,\leq\,   \, &\delta\, C \left( \mathcal{J}_0  + \mathcal{J}_1 \right) +  \, \delta^2  \,C\exp \left( \, \mathcal{H}_0  \left( \mathcal{J}_0  + \mathcal{J}_1 \right) \right)   +  \delta^2  \,C \exp \left( \, \mathcal{H}_1  \left( \mathcal{J}_0  + \mathcal{J}_1 \right) \right) + \mathcal{O}(\delta^3),
\end{align*} 
where $C$ is a constant dependent on T. $\mathcal{J}_1$ and $\mathcal{J}_2$ are Jacobian-related terms, and $\mathcal{H}_1$ and $\mathcal{H}_2$ are Hessian-related terms.
\end{theorem}
The Jacobian-related terms $\mathcal{J}_1$ and  $\mathcal{J}_2$ are defined as $\mathcal{J}_0 :=  \exp \left( \mathcal{F}_h + \mathcal{F}_z + \mathcal{P}_h \right), \, \mathcal{J}_1 := \exp \left( \bar{\mathcal{P}}_h \right)$; the Hessian-related terms $\mathcal{H}_0$ and $\mathcal{H}_1$ are defined as $\mathcal{H}_0 :=  \mathcal{F}_{hh} + \mathcal{F}_{hz} + \mathcal{F}_{zh}+ \mathcal{F}_{zz} +  \mathcal{P}_{hh}, \mathcal{H}_1 := \bar{\mathcal{P}}_{hh}$, 
where $\mathcal{F}_h$, $\mathcal{F}_z$ are the expected $\sup$ Frobenius norm of Jacobians of $f$ w.r.t $h$, $z$, respectively, and $\mathcal{F}_{hh}, \mathcal{F}_{hz}, \mathcal{F}_{zh}, \mathcal{F}_{zz}$ are the corresponding expected $\sup$ Frobenius norm of second-order derivatives. Other terms are similarly defined. A detailed description of all terms, can be found in Appendix \ref{error-accu}. 

Theorem \ref{acc-thm} correlates with the empirical findings in \cite{hafner2019learning} regarding the diminished predictive accuracy of latent states $\tilde{z}_t$ over the extended horizons. In particular, Theorem \ref{acc-thm} suggests that the expected divergence from error accumulation hinges on the expected error magnitude, the Jacobian norms within the latent dynamics model and the horizon length $T$.  

Our next result reveals how initial latent representation error influences the value function $Q$ during the prediction rollouts, which again verifies that the perturbation is dependent on expected error magnitude, the model's Jacobian norms and the horizon length $T$:
\begin{corollary}\label{acc-val-thm}
For a square-integrable $\varepsilon$, let $x_t := (h_t, z_t)$. Then, for any action $a \in \mathcal{A}$, the following holds for value function $Q$ almost surely: 
\begin{align*}
    Q(x_t^{\varepsilon}, a) = & \,Q(x_t^0, a) + \frac{\partial}{\partial x} Q(x_t^0, a) \left( \varepsilon^i \partial_i \, x_t^0 +\frac{1}{2}\varepsilon^i\, \varepsilon^j \,\partial^2_{ij}\,x_t^0  \right)\\& + \frac{1}{2}(\varepsilon^i\, \partial_i\, x_t^0)^\top \frac{\partial^2}{\partial x^2} Q(x_t^0, a) \,(\varepsilon^i\, \partial_i\, x_t^0) +\mathcal{O}(\delta^3),
\end{align*}
as  $\delta \rightarrow 0$, where stochastic processes $\partial_i \, x_t^0$, $\partial^2_{ij} \, x_t^0$ are the first and second derivatives of $x_t^0$ w.r.t $\varepsilon$ and are bounded as follows: 
\begin{equation*}
\E \sup_{t \in [0, T]} \left\|\partial_i \, x_t^0\right\| \leq  C \left( \mathcal{J}_0  + \mathcal{J}_1 \right), \, \E \sup_{t \in [0,T]}  \left\|\partial^2_{ij} \, x_t^0\right\| \leq C \exp \left( \, \mathcal{H}_0  \left( \mathcal{J}_0  + \mathcal{J}_1 \right) \right)   +C \exp \left( \, \mathcal{H}_1  \left( \mathcal{J}_0  + \mathcal{J}_1 \right) \right) .
\end{equation*}
\end{corollary}
This corollary reveals that latent representation errors implicitly encourage exploration of unseen states by inducing a stochastic perturbation in the value function, which again can be regularized through a controlled Jacobian norm. Intuitively, the stochasticity in the LDM also encourages greater exploration compared to its deterministic counterparts.




\textbf{Jacobian Regularization against Non-Zero Drift. \label{jac_reg_dis}}
The above theoretical results have established a close connection of input-output Jacobian matrices with the stabilized generalization capacity of world models (shown in \ref{fund-ineq} under non-zero drift form), and perturbation magnitude in predictive rollouts (indicated in the presence of Jacobian terms in Theorem \ref{acc-thm} and Corollary \ref{acc-val-thm}.) Building on these insights,  we propose a regularizer on input-output Jacobian norm  $\|\frac{\partial g_k}{\partial x}\|_F$ that could modulate $\tilde{\xi}$  ( and in addition $\xi_k$). This regularization not only enhances robustness by controlling perturbations but also reinforces generalization through smoother dynamics in the world model's latent space.  

The regularized loss function for LDM is defined as follows:
\begin{equation}
    \bar{\mathcal{L}}_{\text{dyn}} = \mathcal{L}_{\text{dyn}} + \lambda \, \|J_\theta\|_F,  
\end{equation}
where $\mathcal{L}_{\text{dyn}}$ is the original loss function for dynamics model, $J_\theta$ denotes the data-dependent Jacobian matrix associated with the $\theta$-parameterized dynamics model, and $\lambda$ is the regularization weight (usually chosen from range $[0.01, 0.1]$, see \cite{hoffman2019robust}). Our empirical results in \ref{gen-exp} with an emphasis on sequential case align with the experimental findings from \cite{hoffman2019robust} that Jacobian regularization can enhance robustness against random and adversarial input perturbation.

\section{Experimental Studies}

In this section, extensive experiments are carried out over a number of  tasks in Mujoco environments. Due to space limitation, implementation details and additional results, including the standard deviation of the trials, are relegated to Section \ref{exp-appendix} in the Appendix. 

\noindent {\bf Enhanced robustness and generalization to unseen noisy states and varied dynamics.\label{gen-exp}}We evaluated the effectiveness of Jacobian regularization by comparing a model trained with this regularization against a vanilla model during inference, using perturbed state images and varied dynamics. We consider three types of \textit{perturbations to the observations}: (1) Gaussian noise applied across the entire image, denoted as $\mathcal{N}(\mu_1, \sigma_1^2)$; (2) rotation; and (3) Gaussian noise applied to a random portion of the image, $\mathcal{N}(\mu_2, \sigma_2^2)$. Additionally, we examine variations in the gravity constant $g$ for \textit{unseen dynamics}. These perturbation patterns align with those commonly used in robustness studies (\cite{curi2021combining, sun2023exploring, zhou2023natural}).

For the Walker task, the parameters are set as $\mu_1 = \mu_2 = 0.5$ and $\sigma_2^2 = 0.15$, while for the Quadruped task, $\mu_1 = 0$, $\mu_2 = 0.05$, and $\sigma_2^2 = 0.2$. In each case, we investigate a range of noise levels: (1) variance $\sigma^2$ ranging from $0.05$ to $0.55$; (2) rotation angles $\alpha$ of $20^\circ$ and $30^\circ$; and (3) masked image percentages $\beta\%$ ranging from $25\%$ to $75\%$. For the unseen dynamics, the gravity constant $g$ is varied from $9.8$ to $1$.

\begin{figure}[!htb]
    \centering
    \includegraphics[width=1.0\linewidth]{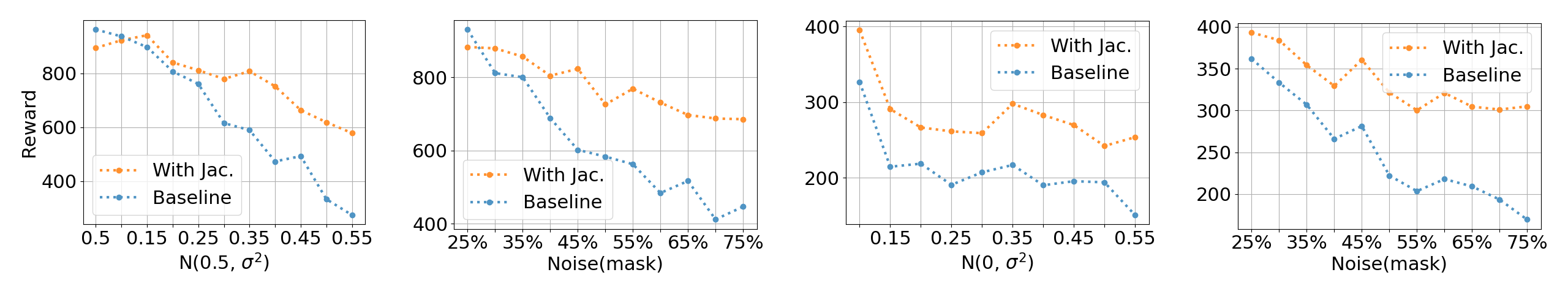}
    \caption{Generalization against increasing degree of perturbation.}
    \label{fig:gen_plot}
\end{figure}

It can be seen from Table \ref{table:expr1-table-rob} and Figure \ref{fig:gen_plot} that thanks to the adoption of Jacobian regularization in training,
the  rewards (averaged over 5 trials) are higher compared to the baseline, indicating improved
robustness to noisy image states in all cases. Moreover, Table \ref{table:expr1-table-rob}demonstrates that the model trained with Jacobian regularization consistently outperforms the baseline under most dynamics variations. These experimental results support the findings in Corollary \ref{unstable-imp-thm}, showing that regularizing the Jacobian norm effectively stabilizes the implicit regularization process, leading to enhanced performance and robustness.

\begin{table}[!htb]
{\footnotesize
\centering
\begin{tabular}{c|c|cc|cc|cc}
\toprule
 &   &\multicolumn{2}{c}{full, $\mathcal{N}(\mu_1, \,\sigma_1^2)$}  & \multicolumn{2}{c}{rotation, $+\alpha^{\circ}$} & \multicolumn{2}{c}{mask $\beta\%$, $\mathcal{N}(\mu_2, \,\sigma_2^2)$} \\
\hline
 & clean & $\sigma_1^2 = 0.35$  & $\sigma_1^2 = 0.5$   &$\alpha=20$& $\alpha=30$& $\beta = 50$ & $\beta =75$\\
\hline
Jac Reg (Walker) & \textbf{967.12}         &    \textbf{742.32}&       \textbf{618.98}&    \textbf{ 423.81}&     \textbf{ 226.04}&     725.81& \textbf{  685.49 }\\
Baseline (Walker)  & 966.53 &      615.79&   333.47&       391.65&    197.53&               583.41&    446.74\\
Jac Reg (Quad) & \textbf{971.98}&    \textbf{269.78}&       \textbf{242.15}&    \textbf{787.63}&     \textbf{610.53}&     \textbf{   321.55}& \textbf{  304.92}\\
Baseline (Quad)  & 967.91&      207.33&   194.08&       681.03&    389.41&               222.22&    169.58\\
\bottomrule
\end{tabular}
\caption{Evaluation on unseen states by various perturbation (Clean means without perturbation). $\lambda=0.01$.}
\label{table:expr1-table-rob}
\vspace{-14 pt}
}
\end{table}

\begin{table}[!htb]
{\footnotesize
\centering
\begin{tabular}{c|c|c|c|c}
\toprule
 &   g = 9.8 &g = 6  & g = 4 & g = 2 \\
\hline
Jac Reg (Walker) & \textbf{967.12}         &    \textbf{906.42}&       \textbf{755.18}&    \textbf{ 679.24}\\
Baseline (Walker)  & 966.53 &      750.36 &   662.86&       381.14\\
Jac Reg (Quad) & \textbf{971.98}&    752.7&       \textbf{543.44}&    \textbf{400.94}\\
Baseline (Quad)  & 967.91&   \textbf{875.02}   & 518.7  &   329.06    \\
\bottomrule
\end{tabular}
\caption{Evaluation on unseen dynamics by various gravity constants ($g = 9.8$ is default). $\lambda=0.01$.}
\label{table:expr1-table-gen}
\vspace{-14 pt}
}
\end{table}

In  some cases where additional knowledge about perturbation is available, such as when the perturbation type is known a priori (which could be unrealistic), one could consider using augmentation methods by training with perturbed observations to improve robustness. We provide a comparative discussion between Jacobian regularization and augmentation methods in the Appendix \ref{appendix-aug}.

\noindent {\bf Robustness against encoder errors.} 
Next, we examine the effects of Jacobian regularization on controlling the error process of the latent states $z$ during training.  Since explicitly characterizing latent representation errors and their drift is impractical, we consider to evaluate the robustness against two exogenous error signals, namely  (1) zero-drift error with  $\mu_t = 0, \sigma_t^2$ ($\sigma_t^2=5$ in Walker, $\sigma_t^2=0.1$ in Quadruped), and (2) non-zero-drift error with $\mu_t \sim [0, 5], \sigma_t^2 \sim [0,5]$ uniformly.  Table \ref{exp_table_inj_error} shows that the model with regularization can consistently learn policies with high returns  and also converges faster, compared to the vanilla case. This corroborates our theoretical findings in Corollary \ref{unstable-imp-thm} that the impacts of error to loss $\mathcal{L}$ can be controlled through the model’s Jacobian norm.

\begin{table}[!htb]
{\footnotesize
\centering
\begin{tabular}{c|cc|cc|cc|cc}
\toprule
 & \multicolumn{2}{c}{Zero drift, Walker} & \multicolumn{2}{c}{Non-zero drift, Walker} &  \multicolumn{2}{c}{Zero drift, Quad} & \multicolumn{2}{c}{Non-zero drift, Quad} \\
\hline
 & 300k & 600k & 300k & 600k & 600k& 1.2M& 1M & 2M \\
\hline
Jac Reg &\textbf{666.2} & \textbf{966} & \textbf{905.7} & \textbf{912.4} & \textbf{439.8}& \textbf{889}& \textbf{348.3} & \textbf{958.7} \\
Baseline & 24.5 & 43.1 & 404.6& 495&   293.6&          475.9&48.98& 32.87\\
\bottomrule
\end{tabular}
\caption{Accumulated rewards under additional encoder errors. $\lambda=0.01$.\label{exp_table_inj_error}}
\label{table:expr1-table-gen}
\vspace{-12 pt}
}
\end{table}

To observe the error propagation of zero-drift and non-zero-drift error signals in latent states, we refer to the visualizations of reconstructed state trajectory samples in the Appendix \ref{app-vis}.

\noindent {\bf Faster convergence on tasks with extended horizon.}
We further evaluate the efficacy of Jacobian regularization in tasks with extended horizon, particularly by extending the horizon length in MuJoCo Walker from $50$ to $100$ steps. Table \ref{exp_table_extend_horizon} shows that the model with regularization converges significantly faster ($\sim$ 100K steps) than the case without Jacobian regularization  in training. This corroborates results in Theorem \ref{acc-thm} that regularizing the Jacobian norm can reduce error propagation.

\begin{table}[!htb]
\centering
\footnotesize
\begin{tabular}{c|ccc}
\toprule
Num Steps & 100k & 200k & 280k \\
\midrule
Jac Reg ($\lambda = 0.05$) & \textbf{639.1} & \textbf{936.3} & 911.1 \\
Jac Reg ($\lambda = 0.1$)  & 537.5 & 762.6 & \textbf{927.7} \\
Baseline                  & 582.3 & 571.2 & 886.6 \\
\bottomrule
\end{tabular}
\caption{Accumulated rewards for the Walker task with an extended horizon (100 steps, increased from the original 50 steps).}
\label{exp_table_extend_horizon}
\vspace{-12pt}
\end{table}

\section{Conclusion}
In this study, we investigate  the robustness and generalization of world models. 
We  develop an SDE formulation by treating LDM as a stochastic dynamical system, and characterize the effects of latent representation errors for zero-drift and non-zero drift cases. Our findings, based on both theoretic and experimental studies,  reveal that for the case with zero drift, modest latent representation errors can paradoxically  function as implicit regularization and hence   enhance robustness. To mitigate the compounding effects of non-zero drift, we applied Jacobian regularization, which enhanced training stability and robustness. Our empirical studies corroborate that Jacobian regularization improves generalization, broadening the model's applicability in complex environments. This work has the potential to improve the robustness and reliability of RL agents, especially in safety-critical applications like autonomous driving.  Future work can extend this study to other world models such as with transformers-based LDM.
\newpage
\bibliography{hang_cite,cite_nips}

\newpage
\appendix

\begin{center}
  \fontsize{14pt}{17pt}\selectfont 
  \textbf{Supplementary Materials}
\end{center}
In this appendix, we provide the supplementary materials supporting the findings of the main paper on the latent representation of latent representations in world models. The organization is as follows:
\begin{itemize}
    \item In Section \ref{appendix-approx}, we provide proof on showing the approximation capacity of CNN encoder-decoder architecture  in latent representation of world models. 
    \item In Section \ref{explicit reg}, we provide proof on implicit regularization of zero-drift errors and additional effects of non-zero-drift errors by showing a proposition on the general form. 
    \item In Section \ref{error-pred}, we provide proof on showing the effects of non-zero-drift errors during predictive rollouts by again showing a result on the general form.
    \item In Section \ref{exp-appendix}, we provide additional results and implementation details on our empirical studies.
\end{itemize}

\newpage
\section{Approximation Power of Latent Representation with CNN Encoder and Decoder \label{appendix-approx}} \label{appendix-approx}
In this section, we show that the latent representation error, in the form of approximation error corresponding to the widely used CNN encoder-decoder, could be made sufficiently small by finding appropriate CNN network configuration. In particular, this result provides theoretical justification to interpreting latent representation error as stochastic perturbation in the dynamical system defined in Equations (\ref{per-encoder} - \ref{unper-est-s}), as the error magnitude $\varepsilon$ can be made sufficiently small by CNN network configuration, and the analysis carries over to other architectures (e.g., ReLU) along the same line.

 To mathematically describe this \textit{intrinsic lower-dimensional geometric structure}, for an integer $k>0$ and $\alpha \in (0,1]$, we consider the notion of smooth manifold (in the $\mathcal{C}^{k,\alpha}$ sense), formally defined by 
\begin{definition}[$\mathcal{C}^{k,\alpha}$ manifold]
A $\mathcal{C}^{k,\alpha}$ manifold $\mathcal{M}$ of dimension $n$ is a topological manifold (i.e. a topological space that is locally Euclidean, with countable basis, and Hausdorff) that has a $\mathcal{C}^{k,\alpha}$ structure $\Xi$ that is a collection of coordinate charts $\{U_{\alpha}, \psi_{\alpha}\}_{\alpha \in A}$ where $U_{\alpha}$ is an open subset of $\mathcal{M}$, $\psi_{\alpha}: U_{\alpha} \to V_{\alpha} \subseteq \mathbb{R}^n$ such that 
\begin{itemize}
    \item  $ \bigcup_{\alpha \in A} U_{\alpha} \supseteq \mathcal{M}$, meaning that the the open subsets form an open cover, 
    \item Each chart $\psi_\alpha$ is a diffeomorphism that is a smooth map with smooth inverse (in the $\mathcal{C}^{k,\alpha}$ sense),
    \item Any two charts are $\mathcal{C}^{k,\alpha}$-compatible with each other, that is for all $\alpha_1, \alpha_2 \in A$, $\psi_{\alpha_1} \circ \psi_{\alpha_2}^{-1}: \psi_{\alpha_2} (U_{\alpha_1} \cap U_{\alpha_2}) \to \psi_{\alpha_1} (U_{\alpha_1} \cap U_{\alpha_2})$ is $\mathcal{C}^{k,\alpha}$.
\end{itemize}
\end{definition}
Intuitively, a $\mathcal{C}^{k,\alpha}$ manifold is a generalization of Euclidean space by allowing additional spaces with nontrivial global structures through a collection of charts that are diffeomorphisms mapping open subsets from the manifold to open subsets of euclidean space.  For technical utility, the defined charts allow to transfer most familiar real analysis tools to the manifold space. For more references, see \cite{Lee2018}.

\begin{definition}
[Riemannian volume form]
Let $\mathcal{X}$ be a smooth, oriented $d$-dimensional manifold with Riemannian metric $g$. A volume form $d \text{vol}_\mathcal{M}$ is the canonical volume form on $\mathcal{X}$ if for any point $x \in \mathcal{X}$, for a chosen local coordinate chart $(x_1, ..., x_d)$, $d \text{vol}_{\mathcal{M}} = \sqrt{\det g_{ij}}\, dx_1 \,\wedge \,... \,\wedge\, dx_d$, where $g_{ij}(x) := g\,(\frac{\partial}{\partial x_i}, \frac{\partial}{\partial x_j})(x)$.  
\end{definition}
Then the induced volume measure by the canonical volume form $d\text{vol}_\mathcal{X}$ is denoted as  $\mu_{\mathcal{X}}$ , defined by $\mu_\mathcal{X}:  \, A  \mapsto \int_A d\text{vol}_\mathcal{X}$, for any Borel-measurable subset $A$ on the space $\mathcal{X}$.  For more references, see \cite{Evans2015-cv}.

We recall the latent representation problem defined in the main paper.

Consider the state space $\mathcal{S} \subset \R^{d_\mathcal{S}}$ and the latent space $\mathcal{Z}$. Consider a state probability measure $Q$ on the state space $\mathcal{S}$ and a probability measure $P$ on the latent space $\mathcal{Z}$. 

\begin{assumption} (Latent manifold assumption)
For a positive integer $k$, there exists a $d_{\mathcal{M}}$-dimensional $\mathcal{C}^{k, \alpha}$ submanifold $\mathcal{M}$ (with $\mathcal{C}^{k+3, \alpha}$ boundary) with Riemannian metric $g$ and has positive reach and also isometrically embedded in the state space $\mathcal{S} \subset \R^{d_\mathcal{S}}$ and $d_{\mathcal{M}} \,<\!\!< d_\mathcal{S}$, where the state probability measure is supported on. In addition, $\mathcal{M}$ is a compact, orientable, connected manifold.   \label{mfld ass-appendix}
\end{assumption}

\begin{assumption} (Smoothness of state probability measure)
$Q$ is a probability measure supported on $\mathcal{M}$ with its Radon-Nikodym derivative $q \in \mathcal{C}^{k,\alpha}(\mathcal{M}, \R)$ w.r.t $\mu_{\mathcal{M}}$.  \label{density smooth ass-appendix}
\end{assumption}

Let $\mathcal{Z}$ be a closed ball in $\R^{d_{\mathcal{M}}}$, that is $\{ x \in \R^ {d_{\mathcal{M}}} \,:\, \|x\| \leq 1 \,\}$.
$P$ is a probability measure supported on $\mathcal{Z}$ with its Radon-Nikodym derivative $p \in \mathcal{C}^{k,\alpha}(\mathcal{Z}, \R)$ w.r.t $\mu_{\mathcal{Z}}$.

We consider a real-valued CNN function \( f_{\text{CNN}} : \mathcal{X} \to \mathbb{R} \), as it can be easily extended to the definition in the $\mathbb{R}^n$-valued case. Let \( f_{\text{CNN}} \) have \( L \) hidden layers, represented as:
\[
f_{\text{CNN}}(x) = A_{L+1} \circ A_L \circ \cdots \circ A_2 \circ A_1(x), \quad x \in \mathcal{X},
\]
where \( A_i \)'s are either convolutional or downsampling operators. For convolutional layers, 
\[
A_i(x) = \sigma(W_i^c x + b_i^c),
\]
where \( W_i^c \in \mathbb{R}^{d_i \times d_{i-1}} \) is a structured sparse Toeplitz matrix from the convolutional filter \(\{w_j^{(i)}\}_{j=0}^{s(i)}\) with filter length \( s(i) \in \mathbb{N}_+ \), \( b_i^c \in \mathbb{R}^{d_i} \) is a bias vector, and \(\sigma\) is the ReLU activation function. 

For downsampling layers,
\[
A_i(x) = D_i(x) = \left(x_{j m_i} \right)_{j=1}^{\lfloor d_{i-1}/m_i \rfloor},
\]
where \( D_i : \mathbb{R}^{d_i \times d_{i-1}} \) is the downsampling operator with scaling parameter \( m_i \leq d_{i-1} \) in the \(i\)-th layer. The convolutional and downsampling operations are elaborated in Appendix [63]. We examine the class of functions represented by CNNs, denoted by \( \mathcal{F}_{\text{CNN}} \), defined as:
\[
\mathcal{F}_{\text{CNN}} = \{ f_{\text{CNN}} \text{ as in defined above with any choice of } A_i, \, i = 1, \ldots, L+1 \}.
\]
For more details in the definitions of CNN functions, we refer to \cite{shen2022approximation}.

\begin{assumption}
Assume that $\mathcal{M}$ and $\mathcal{Z}$ are locally diffeomorphic, that is there exists a map $F: \mathcal{M} \rightarrow \mathcal{Z}$ such that at every point $x$ on $\mathcal{M}$, $\, \det(d\,F(x)) \neq 0$.\label{local diffeo ass-appendix}
\end{assumption}

\begin{theorem}
\textbf{(Approximation Error of Latent Representation).}  \label{approx-thm-main-appendix}
Under Assumption \ref{mfld ass-appendix}, \ref{density smooth ass-appendix} and \ref{local diffeo ass-appendix}, for $\theta \in (0, 1),$ let $d_\theta = \mathcal{O}(d_\mathcal{M}\theta^{-2}\log \frac{d}{\theta})$. For  positive integers $M$ and $N$, there exists an encoder $g_\text{enc}$ and decoder $g_\text{dec}$ $\in \mathcal{F}_\text{CNN}(L, S, W)$ s.t. 
\begin{align*}
&W_1(g_{\text{enc}_\#} Q, P) \leq d_\mathcal{M} C (NM)^{-\frac{2(k+1)}{d_\theta}},\\
&W_1(g_{\text{dec}_\#} P, Q) \leq d_\mathcal{M} C (NM)^{-\frac{2(k+1)}{d_\theta}}.
\end{align*}
\end{theorem}
The primary challenge to show Theorem  \ref{approx-thm-main-appendix} is in demonstrating the existence of oracle encoder and decoder maps. These maps, denoted as $g_\text{enc}^*: \mathcal{M} \rightarrow \mathcal{Z}$ and $g_\text{dec}^*: \mathcal{Z} \rightarrow \mathcal{M}$ respectively, must satisfy  
\begin{equation}
        {g_\text{enc}^*}_{\#}\,Q = P, \quad {g_\text{dec}^*}_{\#}\,P = Q.
\end{equation}
and importantly they have the proper smoothness guarantee, namely $g_\text{enc}^* \in \mathcal{C}^{k+1,\alpha}(\mathcal{M}, \mathcal{Z})$ and $g_\text{dec}^* \in \mathcal{C}^{k+1,\alpha}(\mathcal{Z}, \mathcal{M})$.  Proposition \ref{mfd-fun} shows the existence of such oracle map(s).

\begin{proposition} [$\mathcal{C}^{k, \alpha}$, compact]\label{mfd-fun}
\label{exist-oracle}
Let $\mathcal{M}, \mathcal{N}$ be compact, oriented d-dimensional Riemannian manifolds with $\mathcal{C}^{k+3, \alpha}$ boundary with the volume measure $\mu_{\mathcal{M}}$ and $\mu_{\mathcal{N}}$ respectively. Let $Q, \,P$ be distributions supported on $\mathcal{M}, \, \mathcal{N}$ respectively with their $\mathcal{C}^{k,\alpha}$ density functions $q, \,p$, that is $Q, \,P$ are probability measures supported on $\mathcal{M}, \, \mathcal{N}$ with their Radon-Nikodym derivatives $q \in \mathcal{C}^{k,\alpha}(\mathcal{M}, \R)$ w.r.t $\mu_{\mathcal{M}}$ and $p \in \mathcal{C}^{k,\alpha}(\mathcal{N}, \R)$ w.r.t $\mu_{\mathcal{N}}$. Then, there exists a $\mathcal{C}^{k+1,\alpha}$ map $g: \mathcal{N} \rightarrow \mathcal{M}$ such that the pushforward measure $g_{\#}P=Q$, that is for any measurable subset $A \in \mathcal{B}(\mathcal{M})$, $\,Q(A)=P(g^{-1}(A))$.  
\end{proposition}

\begin{figure}
    \centering
    \includegraphics[width=0.8\linewidth]{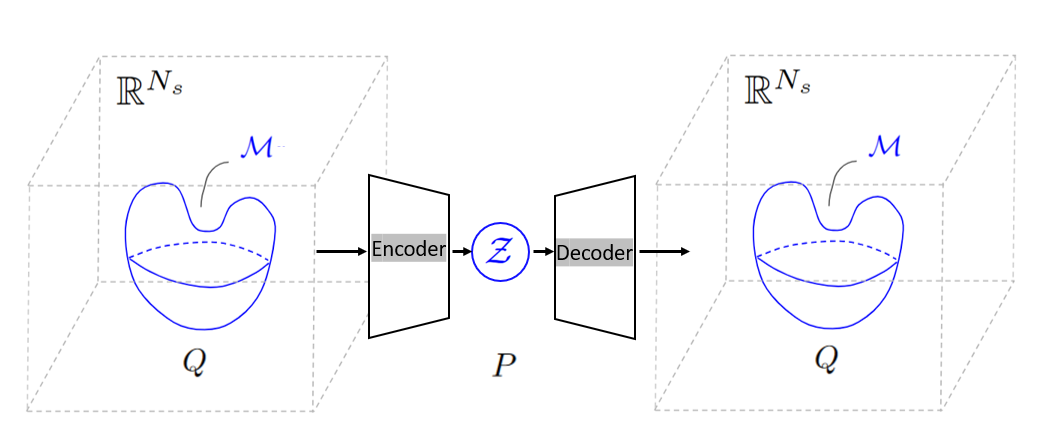}
    \vspace{-15 pt}
    \caption{\textit{Latent Representation Problem}: The left and right denote the manifold $\mathcal{M}$ with lower dim $d_\mathcal{M}$ embedded in a larger Euclidean space, with latent space $Z$ a $d_\mathcal{M}$-dimensional ball in middle. Encoder and decoder as maps respectively pushing forward Q to P and P to Q.}
    \label{fig:enter-label}
    \vspace{-10 pt}
\end{figure}

\begin{proof} (\textit{Proposition \ref{mfd-fun}})
Let $\omega:=p\,d\text{vol}_{\mathcal{N}}$, then $\omega$ is a $\mathcal{C}^{k,\alpha}$ non-vanishing form on $\mathcal{N}$, as $p \in \mathcal{C}^{k,\alpha}$ and for any point $x \in \mathcal{N}$, we have $p(x) > 0$. In addition,
$\int_\mathcal{N} \omega = \int_\mathcal{N} p\,d\text{vol}_{\mathcal{N}} = \int_\mathcal{N} p\,d\mu_{\mathcal{N}} = P(\mathcal{N}) = 1.$ Similarly, let $\eta:=q\,d\text{vol}_{\mathcal{M}}$ a $\mathcal{C}^{k,\alpha}$ non-vanishing form on $\mathcal{M}$ and $\int_\mathcal{M} \eta = 1$.\\

Let $F: \mathcal{N} \rightarrow \mathcal{M}$ be an orientation-preserving local diffeomorphism, we then have $\det( dF ) > 0$ everywhere on $\mathcal{N}$. \\
As $\mathcal{N}$ is compact and $\mathcal{M}$ is connected by assumption, $F$ is a covering map, that is for every point $x \in \mathcal{M}$, there exists an open neighborhood $U_x$ of $x$ and a discrete set $D_x$ such that $F^{-1}(U) = \sqcup_{\alpha\in D} \,V_{\alpha} \subset \mathcal{N}$ and $F|_{V_\alpha} = V_{\alpha} \rightarrow U$ is a diffeomorphism. Furthermore,  $|D_x| = |D_y|$ for any points $x, y \in \mathcal{M}$. In addition, $|D_x|$ is finite from the compactness of $\mathcal{N}$.  

Let $\bar{\eta}$ be the pushforward of $\omega$ via $F$,  defined by for any point $x \in \mathcal{M}$ and a neighborhood $U_x$,
\begin{equation}
    \bar{\eta}(x):= \frac{1}{|D_x|} \sum_{\alpha \in D_x} \left({F\big|_{V_{\alpha}}}^{-1}\right)^{*} \omega\big|_{V_{\alpha}}.
\end{equation}
$\bar{\eta}$ is well-defined as it is not dependent on the choice of neighborhoods and the sum and $\frac{1}{|D_x|}$ are always finite.  Furthermore,  $\bar{\eta}$ is a $\mathcal{C}^{k,\alpha}$ non-vanishing form on $\mathcal{M}$, as $p \circ \left({F\big|_{V_{\alpha}}}^{-1}\right)$ is $\mathcal{C}^{k,\alpha}$. \\

Notice that ${F\big|_{V_{\alpha}}}^{-1}$ is orientation-preserving as $\det d\,{F\big|_{V_{\alpha}}}^{-1} = \frac{1}{\det d\,F\big|_{V_{\alpha}}} > 0$ everywhere on $V_\alpha$. As ${F\big|_{V_{\alpha}}}^{-1}$ is an orientation-preserving diffeomorphism, then its degree $c: = \deg({F\big|_{V_{\alpha}}}^{-1}) = 1$. Then, $\int_{\mathcal{M}}\bar{\eta} = c \int_{\mathcal{N}}\omega = 1$. \\

As we have shown that $\eta$ and $\bar{\eta} \, \in \mathcal{C}^{k,\alpha}$ and $\int_{\mathcal{M}}\bar{\eta} = \int_{\mathcal{M}}\eta$, by \cite{AIHPC_1990}, there exists a diffeomorphism $\psi: \mathcal{M} \rightarrow \mathcal{M}$ fixing on the boundary such that $\psi^{*}\eta = \bar{\eta}$, where $\psi, \psi^{-1} \in \mathcal{C}^{k+1,\alpha}$.\\
Let $g:= \psi \circ F$, then it holds that $g^{*}\eta = (\psi \circ F)^* \eta = F^{*}\circ\psi^{*}\eta = F^{*}\bar{\eta} = \omega $.

Then, for any measurable subset $A$ on the manifold $\mathcal{M}$, we verify that
$Q(A) = \int_A \eta = \int_{g^{-1}(A)} g^*\eta = \int_{g^{-1}(A)} \omega = \int_{g^{-1}(A)} p\,d\text{vol}_{\mathcal{N}} = \int_{g^{-1}(A)} p\,d\mu_{\mathcal{N}} = P(g^{-1}(A))$.\\

Hence, we have shown the existence by an explicit construction. As $\psi \in \mathcal{C}^{k+1,\alpha}$, and $F \in \mathcal{C}^\infty$, then we have $g \in \mathcal{C}^{k+1,\alpha}$. 
\end{proof}

We are now ready to show Theorem \ref{approx-thm-main-appendix} with the existence of oracle map and the low-dimensional approximation results from \cite{shen2022approximation}.  
\begin{proof}
(\textit{Theorem \ref{approx-thm-main-appendix}})
For encoder, from Proposition \ref{exist-oracle}, there exists an $\mathcal{C}^{k+1,\alpha}$ oracle map $g: \mathcal{M} \rightarrow \mathcal{Z}$ such that the pushforward measure $g_{\#}Q=P$.  
Then, 
\begin{align*}
    W_1(({g_\text{enc}})_{\#}Q\,,\, P) = & W_1({(g_\text{enc}})_{\#}Q\,,\, g_{\#}Q) \\
    = &\, \sup _{f \in \text{Lip}_1(\mathcal{Z})} \left| \int_\mathcal{Z} f(y) \,d(({g_\text{enc}})_{\#}Q) - \int_\mathcal{Z} f(y) \,d(g_{\#} Q)\right|\\
    \leq &\, \sup _{f \in \text{Lip}_1(\mathcal{Z})} \int_\mathcal{M} \left| f \circ g_\text{enc} (x) - f \circ g(x) \right|\, dQ\\
    \leq &\, \int_\mathcal{M} \left\| g_\text{enc} (x) - g(x) \right\| \,dQ\\
    \leq &\, d_\mathcal{M} C (NM)^{-\frac{2(k+1)}{d_\theta}},
\end{align*}
where the last inequality follows from the special case $\rho=0$ of Theorem 2.4 in  \cite{shen2022approximation}. 

Similarly, for decoder, from Proposition \ref{exist-oracle}, there exists an $\mathcal{C}^{k+1,\alpha}$ oracle map  $\bar{g}: \mathcal{Z} \rightarrow \mathcal{M}$ such that the pushforward measure $\bar{g}_{\#}P=Q$.
\begin{align*}
    W_1(({g_\text{dec}})_{\#}P\,,\, Q) = & W_1({(g_\text{dec}})_{\#}P\,,\, \bar{g}_{\#}P) \\
    \leq &\, \int_\mathcal{Z} \left\| g_\text{dec} (y) - \bar{g}(y) \right\| \,dP\\
    \leq &\, d_\mathcal{M} C (NM)^{-\frac{2(k+1)}{d_\theta}}.
\end{align*}
\end{proof}

\newpage
\section{Explicit Regularization of Latent Representation Error in World Model Learning} \label{explicit reg}

We recall the SDEs for latent dynamics model defined in the main paper.
Consider a complete, filtered probability space $(\Omega, \,\mathcal{F}, \,\{\mathcal{F}_t\}_{t\in[0, T]}, \,\mathbb{P}\,)$ where independent standard Brownian motions $B^{\text{\,enc}}_t,\, B^{\text{\,pred}}_t, B^{\text{\,seq}}_t, \,B^{\text{\,dec}}_t$ are defined such that ${\mathcal{F}_t}$ is their augmented filtration, and $T \in \mathbb{R}$ as the time length of the task environment. We consider the stochastic dynamics of LDM through the following coupled SDEs after error perturbation: 
\begin{align}
&  d\,z_t =  \left( q_{\text{enc}}(h_t, s_t) + \sigma(h_t, s_t)\right)\,dt  + \left(\bar{q}_{\text{enc}}(h_t, s_t) + \bar{\sigma}(h_t, s_t) \right)d B^\text{\,enc}_t,\label{per-encoder-appendix}\\
& d\,h_t  = f(h_t, z_t, \pi(h_t, z_t)) \, dt +  \bar{f}(h_t, z_t, \pi(h_t, z_t)) \,d B^{\text{\,seq}}_t \label{unper-h-appendix}\\
& d\,\tilde{z}_t = p(h_t)\,dt + \bar{p} (h_t)\,dB^{\text{\,pred}}_t,\label{unper-z-appendix}\\
&  d\,\tilde{s}_t = q_{\text{dec}}(h_t, \tilde{z}_t)\,dt + \bar{q}_{\text{dec}}(h_t, \tilde{z}_t)\,dB^{\text{\,dec}}_t \label{unper-est-s-appendix},
\end{align}
where $\pi(h, \tilde{z})$ is a policy function as a local maximizer of value function and the stochastic process $s_t$ is $\mathcal{F}_t$-adapted. 

As discussed in the main paper, our analysis applies to a common class of world models that uses Gaussian distributions parameterized by neural networks' outputs for $z$, $\tilde{z}$, $\tilde{s}$. Their distributions are not non-Gaussian in general. 

For example, as $z$ is conditional Gaussian and its mean and variance are random variables which are learned by the encoder from r.v.s $s$ and $h$ as inputs, thus rendering $z$ non-Gaussian. However, $z$ is indeed Gaussian when the inputs are known. Under this conditional Gaussian class of world models, to see that the continuous formulation of latent dynamics model can be interrupted as SDEs, one notices that SDEs with coefficient functions of known inputs are indeed Gaussian, matching to this class of world models. Formally, in the context of $z$ without latent representation error:  
\begin{proposition} (Latent states SDE conditioned on inputs is Gaussian) \label{prop-lat-gau}\\
For the latent state process $z_{t\in[0,T]}$ without error,
\begin{equation}
    d\,z_t =   q_{\text{enc}}(h_t, s_t) \,dt  + \bar{q}_{\text{enc}}(h_t, s_t))d B^\text{\,enc}_t,\label{encoder-appendix}
\end{equation}
 with zero initial value. Given known $h_{t\in[0,T]}$ and $s_{t\in[0,T]}$, the process $z_t$ is a Gaussian process. Furthermore, for any $t \in [0,T]$, $z_t$ follows a Gaussian distribution with mean $\mu_t = \int_0^t q_{\text{enc}}(h_s, s_s) ds$ and variance $\sigma^2_t = \int_0^t \bar{q}_{\text{enc}}(h_s, s_s)^2 ds$.
\end{proposition}
\begin{proof}
Proof follows from Proposition 7.6 in \cite{Steele_2001}. 
\end{proof}
Next, we recall our assumptions from the main text:
\begin{assumption}\label{reg-sde-appendix}
The drift coefficient functions $q_\text{enc},$ $f,$  $p$ and $q_\text{dec}$ and the diffusion coefficient functions $\bar{q}_\text{enc},  \,\bar{p}$ and $\bar{q}_{\text{dec}}$  are bounded and Borel-measurable over the interval $[0, T]$, and of class $\mathcal{C}^3$ with bounded Lipschitz continuous partial derivatives. The initial values $z_0, h_0, \tilde{z}_0, \tilde{s}_0$ are square-integrable random variables. 
 \end{assumption}
\begin{assumption}
$\sigma$ and $\bar{\sigma}$ are bounded and Borel-measurable and are of class $\mathcal{C}^3$ with bounded Lipschitz continuous partial derivatives over the interval $[0, T]$. \label{noise-reg-appendix}
\end{assumption}

One of our main results is the following:
\begin{theorem} (Explicit Regularization Induced by Zero-Drift Representation Error)\label{imp-reg-thm-appendix}\\
Under Assumption \ref{reg-sde-appendix}  and \ref{noise-reg-appendix} and considering a loss function $\mathcal{L} \in \mathcal{C}^2$, the explicit effects of the zero-drift error can be marginalized out as follows:
\begin{equation}
    \E\, \mathcal{L}\left(x^\varepsilon_t\right) = \E\,\mathcal{L}(x_t^0) +  \mathcal{R}  + \mathcal{O}(\varepsilon^3),
\end{equation}
as $\varepsilon \rightarrow 0$, where the regularization term $\mathcal{R}$ is given by 
$\mathcal{R} := \, \varepsilon\, \mathcal{P} + \varepsilon^2 \left(\mathcal{Q} +  \frac{1}{2}\,\mathcal{S} \right).$\\
Each term of $\mathcal{R}$ is as follows:
\begin{align}
    \mathcal{P} := & \,\E \,\nabla \mathcal{L}(x_t^0)^\top \Phi_t \sum _k \xi_t^k,\\
    \mathcal{Q} := & \,\E \, \nabla \mathcal{L} (x_t^0)^\top  \Phi_t \int_0^t \Phi_s^{-1} \, \mathcal{H}^k(x_s^0, s) dB^k_t, \\
    \mathcal{S} := & \,\E \sum _{k_1, k_2} (\Phi_t \xi_t^{k_1})^i \nabla^2 \mathcal{L}(x_t^0, t)\,(\Phi_t \xi_t^{k_2})^j,
\end{align}
where square matrix $\Phi_t$ is the stochastic fundamental matrix of the corresponding homogeneous equation:
\begin{equation*}
    d \Phi_t = \frac{\partial \bar{g}_k}{\partial x}(x_t^0, t)\, \Phi_t\, dB^k_t, \quad \Phi(0) = I,
\end{equation*}
and $\xi_t^k$ is as the shorthand for $\int_0^t \Phi_s^{-1}\bar{\sigma}_k(x_s^0, s)dB^k_t$. Additionally, $\mathcal{H}^k(x_s^0, s)$ is represented by for $\sum _{k_1, k_2} \frac{\partial^2 \bar{g}_{k}}{\partial x^i \partial x^j}(x^0_s, s)\left( \xi_s^{k_1}\right)^i\left( \xi_s^{k_2}\right)^j$.
\end{theorem}


Before proving Theorem \ref{imp-reg-thm-appendix}, we first show Proposition \ref{main} on the general case of perturbation to the stochastic system. Consider the following perturbed system given by 
\begin{equation}
d\,x_{t}=\left(g_{0}\left(x_{t}, t\right)+\varepsilon \, \eta_{0}\left(x_{t}, t\right)\right) d t+ \sum _{k=1}^m \left(g_{k}\left(x_{t}, t\right)+\varepsilon \, \eta_{k}\left(x_{t}, t\right)\right) d B^k_{t}\label{origSDE}
\end{equation}
with initial values $x(0)=x_{0}$,

\begin{proposition}\label{main}
Suppose that $f$ is a real-valued function that is $\mathcal{C}^{2}$. Then it holds that, with probability 1, as $\varepsilon \rightarrow 0$,  for $t \in[0, T]$,
\begin{equation}\label{taylor-1}
f\left(x_{t}^{\varepsilon}\right)=  f\left(x_{t}^{0}\right)+\varepsilon \nabla f\left(x_{t}^{0}\right)^{\top} \partial_{\varepsilon}\, x_{t}^{0} +\varepsilon^{2}\left(\nabla f\left(x_{t}^{0}\right)^{\top} \partial_{\varepsilon}^{2} x_{t}^{0}\right. \left.+\frac{1}{2} \partial_{\varepsilon}\, {x_{t}^{0}}^{\top} \nabla^2 f\left(x_{t}^{0}\right) \partial_{\varepsilon}\, x_{t}^{0}\right)+\mathcal{O}\left(\varepsilon^{3}\right),
\end{equation}
 where the stochastic process $x^{0}_t$ is the solution to SDE \ref{origSDE} with $\varepsilon=0$, with its first and second-order derivatives w.r.t $\varepsilon$ denoted as $\partial_{\varepsilon}\, x^{0}_t, \partial_{\varepsilon}^{2}\, x^{0}_t$. \\
Furthermore, it holds that $\partial_{\varepsilon}\, x^{0}_t, \partial_{\varepsilon}^{2}\, x^{0}_t$ satisfy the following SDEs with probability 1,
\begin{equation}
\begin{aligned}\label{prop-SDEs}
& d\, \partial_{\varepsilon} x_{t}^{0}=\left(\frac{\partial g_{k}}{\partial x}\left(x_{t}^{0}, t\right) \partial_{\varepsilon} x_{t}^{0}+\eta_{k}\left(x_{t}^{0}, t\right)\right) d B_{t}^{k}, \\
& d\, \partial_{\varepsilon}^{2} x_{t}=\left(\Psi_{k}\left(\partial_{\varepsilon} x_{t}^{0}, x_{t}^{0}, t\right)+2 \frac{\partial \eta_{k}}{\partial x}\left(x_{t}^{0}, t\right) \partial_{\varepsilon} x_{t}^{0}+\frac{\partial g_{k}}{\partial x}\left(x_{t}^{0}, t\right) \partial_{\varepsilon}^{2} x_{t}^{0}\right) d B_{t}^{k} ,\\
\end{aligned}
\end{equation}
with initial values $\partial_{\varepsilon}\, x(0)=0, \partial_{\varepsilon}^{2}\, x(0)=0$, where  
\begin{equation*}
\Psi_{k}:
\left(\partial_{\varepsilon}\, x, x, t\right) \mapsto \partial_{\varepsilon}\, x^{i} \frac{\partial g_{k}}{\partial x^{i} \partial x^{j}}(x, t) \partial_{\varepsilon}\, x^{j},
\end{equation*}
for $k =0, 1, ..., m$.
\end{proposition}

\begin{proof}
We first apply the stochastic version of perturbation theory to SDE \ref{origSDE}. For brevity, we will write $t$ as $B_{t}^{0}$ and use Einstein summation convention. Hence, SDE \ref{origSDE} is rewritten as
\begin{equation} \label{Bt0SDE}
d x_{t}= \gamma_{k}^{\varepsilon}\left(x_{t}, t\right) d B_{t}^{k},
\end{equation}
with initial value $x(0)=x_{0}$.

\textit{Step 1}: We begin with the corresponding systems to derive the SDEs that characterize $\partial_{\varepsilon}\, x_{t}^{\varepsilon}$ and $ \partial_{\varepsilon}^{2}\, x_{t}^{\varepsilon}$. Our main tool is an important result on smoothness of solutions w.r.t. initial data from Theorem 3.1 from Section 2 in \cite{Hennequin_Dudley_Kunita_Ledrappier_1984}. 

For $\partial_{\varepsilon}\, x$, consider the SDEs
\begin{align*} \tag{*} \label{sys4first}
& d\, x_{t}=\gamma_{k}^{\varepsilon}\left(x_{t}, t\right) d B_{t}^{k}, \\
& d\, \varepsilon_{t}=0,
\end{align*}
with initial values $x_{(0)}=x_{0}, \varepsilon(0)=\varepsilon$.
From an application of Theorem 3.1 from Section 2 in \cite{Hennequin_Dudley_Kunita_Ledrappier_1984} on \ref{sys4first}, we have $\partial_{\varepsilon}\, x$  that satisfies the following SDE with probability 1:
\begin{equation}\label{sde1st}
d\, \partial_{\varepsilon} x_{t}=\left(\alpha_{k}^{\varepsilon}\left(x_{t}, t\right) \partial_{\varepsilon} x_{t}+\eta_{k}\left(x_{t}, t\right)\right) d B_{t}^{k} ,
\end{equation}

with initial value $\partial_{\varepsilon} x_{0}=0 \in \mathbb{R}^{n}$, with probability 1, where $x_{t}$ is the solution to  Equation (\ref{Bt0SDE}) and the functions $\alpha_{k}^{\varepsilon}$ are given by
\begin{equation*}
\alpha_{k}^{\varepsilon}: \left(x, t\right) \mapsto \frac{\partial g_{k}}{\partial x^{j}}\left(x, t\right)+\varepsilon \frac{\partial \eta_{k}}{\partial x^{j}}\left(x, t\right),
\end{equation*}
where $k = 0, \,...,\, m$.

To characterize $\partial_{\varepsilon}^{2}\, x_{t}$, consider the following SDEs
\begin{align*}\label{sys4sec}\tag{**}
&d\, x_{t} =\gamma_{k}^{\varepsilon}\left(x_{t}, t\right) d B_{t}^k, \\
&d\, \partial_{\varepsilon}\, x_{t} =\left(\alpha_{k}^{\varepsilon}\left(x_{t}, t\right) \partial_{\varepsilon}\, x_{t}+\eta_{k}\left(x_{t}, t\right)\right) d B_{t}^{k} , \\
&d\, \varepsilon_{t} =0,
\end{align*}
with initial value $x(0)=x_{0}, \,\partial_{\varepsilon}\, x(0)=0, \,\varepsilon(0)=\varepsilon$.

From a similar application of Theorem 3.1 from Section 2 in \cite{Hennequin_Dudley_Kunita_Ledrappier_1984}, the second derivative $\partial_{\varepsilon}^{2}\, x$ satisfies the following SDE with probability 1:

\begin{equation}\label{sde2nd}
 d \,\partial_{\varepsilon}^{2}\, x_{t}=\left( \beta_{k}^{\varepsilon}\left(\partial_{\varepsilon} x_{t}, x_{t}, t\right)+2 \frac{\partial\, \eta_{k}}{\partial x}\left(x_{t}, t\right) \partial_{\varepsilon}\, x_{t} +\alpha_{k}^{\varepsilon}\left(x_{t}, t\right) \partial_{\varepsilon}^{2} x_{t} \right) d B_{t}^{k} ,
\end{equation}

with initial value $\partial_{\varepsilon}^{2}\, x(0)=0 \in \mathbb{R}^{n}$, where $\partial_{\varepsilon}\, x_{t}$ is the solution to Equation(\ref{sde1st}), $x(t)$ is the solution to  Equation (\ref{Bt0SDE}), and the functions
\begin{equation*}
\beta_{k}^{\varepsilon}:
\left(\partial_{\varepsilon}\, x, x, t\right) \mapsto \partial_{\varepsilon}\, x^{j} \left(\frac{\partial g_{k}^{i}}{\partial x^{l} \partial x^{j}}(x,t)+\varepsilon \frac{\partial \eta_{k}^{i}}{\partial x^{l} \partial x^{j}}(x,t)\right) \partial_{\varepsilon}\, x^{l},
\end{equation*}
where $k = 0, \,..., \,m$.

When $\varepsilon=0$ in the obtained SDEs (\ref{Bt0SDE}), (\ref{sde1st}) and (\ref{sde2nd}), the corresponding solutions of which are $x_{t}^{0}, \partial_{\varepsilon}\, x_{t}^{0}, \partial_{\varepsilon}^{2}\, x_{t}^{0}$, we now have the following:
\begin{align}
& d\, x_{t}^{0}= g_{k}\left(x_{t}^{0}, t\right) d B_{t}^{k}, \label{9a}\\
& d\, \partial_{\varepsilon}\, x_{t}^{0}=\left(\frac{\partial g_{k}}{\partial x}\left(x_{t}^{0}, t\right) \partial_{\varepsilon}\, x^{0}+\eta_{k}\left(x_{t}^{0}, t\right)\right) d B_{t}^{k},  \label{9b}\\
& d\, \partial_{\varepsilon}^{2}\, x_{t}^{0}=\left(\Psi_{k}\left(\partial_{\varepsilon}\, x_{t}^{0}, x_{t}^{0}, t\right)+2 \frac{\partial \eta_{k}}{\partial x}\left(x_{t}^{0}, t\right) \partial_{\varepsilon}\,x_{t}^{0}+\frac{\partial g_{k}}{\partial x}\left(x_{t}^{0}, t\right) \partial_{\varepsilon}^{2}\, x_{t}^{0}\right) d B_{t}^{k},  \label{9c}
\end{align}
with initial values $x(0)=x_{0}, \partial_{\varepsilon}\, x(0)=0, \partial_{\varepsilon}^{2}\, x(0)=0$. In particular, $\Psi_{k} := \beta_k^{0}$ is given by

\begin{equation*}
\left(\partial_{\varepsilon} x, x, t\right) \mapsto \partial_{\varepsilon} x^{i} \frac{\partial g_{k}}{\partial x^{i} \partial x^{i}}(x, t) \partial_{\varepsilon} x^{j}.
\end{equation*}

\textit{Step 2}: For the next step, we show that the solutions $x_{t}^{0}, \partial_{s}\, x_{t}^{0}, \partial_{\varepsilon}^{2}\, x_{t}^{0}$ are indeed bounded by proving the following lemma \ref{lem-bd}:

\begin{lemma} \label{lem-bd} 
$$
\mathbb{E} \sup _{t \in[0, T]}\left\|x_{t}^{0}\right\|^{2},\, \mathbb{E} \sup _{t \in[0, T]}\left\|\partial_{\varepsilon}\, x_{t}^{0}\right\|^{2},\text {and  } \mathbb{E} \sup _{t \in[0, T]}\left\|\partial_{\varepsilon}^{2}\, x_{t}^{0}\right\|^{2}
\text { are bounded. }
$$
\end{lemma}
\begin{proof}
To simplify the notations, we take the liberty to write constants as $C$ and notice that $C$ is not necessarily identical in its each appearance.

(1) We first show that $\mathbb{E} \sup _{t \in[0, T]}\left\|x_{t}^{0}\right\|^{2}$ is bounded.

From Equation (\ref{9a}), we have that 
$$
x_{t}^{0}=x_{0}+\int_{0}^{t} g_{k}\left(x_{\tau}, \tau \right) d B_{\tau}^{k} .
$$

By Jensen's inequality. it holds that
\begin{equation}\label{1start}
\mathbb{E} \sup _{t \in[0, T]}\left\|x_{t}\right\|^{2} \leq C\, \E \left\|x_{0}\right\|^{2}+C\, \mathbb{E} \sup _{t \in[0, T]}\left\|\int_{0}^{t} g_{k}\left(x^{0}_{\tau}, \tau\right) d B_{\tau}^{k}\right\|^{2}.
\end{equation}
For the second term on the right hand side, it is a sum over $k$ from $0$ to $m$ by Einstein notation.

For $k=0$, recall that we write $t$ as $B_{t}^{0}$ :
\begin{align*}
 \mathbb{E} \sup _{t \in[0, T]}\left\|\int_{0}^{t} g_{0}\left(x_{\tau}^{0}, \tau\right) d \tau\right\|^{2} \leq & \,C\, \mathbb{E} \sup _{t \in[0, T]} t \int_{0}^{t}\left\|g_{0}\left(x_{\tau}^{0}, \tau\right)\right\|^{2} d \tau, \tag{i}\\
\leq &\, C\, \mathbb{E} \sup _{t \in[0, T]} \int_{0}^{t} C\left(1+\left\|x_{\tau}^{0}\right\|\right)^{2} d \tau, \tag{ii} \\
\leq &\, C\, + C \int_{0}^{T} \mathbb{E} \sup _{s \in[0, \tau]}\left\|x_{s}^{0}\right\|^{2} d \tau,  \tag{iii}
\end{align*}
where we used Jensen's inequality, the assumption on the linear growth, the inequality property of $\sup$ and Fubini's theorem, respectively. 

For $k$ is equal to $1, \dots, m$,
\begin{align*}
\mathbb{E} \sup _{t \in[0, T]}\left\|\int_{0}^{t} g_{1}\left(x_{\tau, \tau}^{0}, \tau\right) d B_{\tau}\right\|^{2} \leq &\,C\, \mathbb{E} \int_{0}^{T}\left\|g_{1}\left(x_{\tau}^{0}, \tau\right)\right\|^{2} d \tau, \tag{iv}\\
\leq & \,C\,+C \int_{0}^{T} \mathbb{E} \sup _{s \in[0, \tau]}\left\|x_{s}^{0}\right\| d \tau , \tag{v}
\end{align*}
where (iv) holds from the Burkholder-Davis-Gundy inequality as $\int_{0}^{t} g_{k}\left(x_{\tau}^{0}, \tau\right) d B_{\tau}$ is a continuous local martingale with respect to the filtration $\mathcal{F}_{t}$; and then one can obtain (v) by following a similar reasoning of (ii) and (iii). 

Hence, now from the previous inequality (\ref{1start}),
$$
\mathbb{E} \sup _{t \in[0, T]}\left\|x_{t}^{0}\right\|^{2} \leq \, \E \left\|x_{0}\right\|^{2}+C+C \int_{0}^{T} \mathbb{E} \sup _{s \in[0, \tau]}\left\|x_{s}^{0}\right\| d \tau .
$$

By  Gronwall's Lemma, it holds true that
$$
\mathbb{E} \sup _{t \in[0, T]}\left\|x_{t}^{0}\right\|^{2} \leq \left(C\,\E\left\|x_{0}\right\|^{2}+C \right) \exp (C ) .
$$
As $x_{0}$ is square-integrable by assumption, therefore we have shown that $\mathbb{E} \sup _{t \in[0, T]}\left\|x_{t}^{0}\right\|^{2}$ is bounded.

(2) We then show that $\E\supp{t\in[0,T]}||\partial_\varepsilon\, x^0_t||^2$ is also bounded.

From the SDE (\ref{9b}), as we have derived that
$$
\partial_{\varepsilon}\, x^{0}_t=\int_{0}^{t} \frac{\partial g_{k}}{\partial x}\left(x_{\tau}^{0}, \tau\right) \partial_{\varepsilon}\, x_{\tau}^{0}+\eta_{k}\left(x_{\tau}^{0}, \tau\right) d B_{\tau}^{k},
$$
then we have
$$
\begin{aligned}
\mathbb{E} \sup _{t \in[0, \tau]}\left\|\partial_{\varepsilon}\, x_{t}^{0}\right\|^{2} \leq \,C\, \mathbb{E} \sup _{t \in[0, \tau]} \left\|\int_{0}^{t} \frac{\partial g_{k}}{\partial x}\left(x_{\tau}^{0}, \tau\right) \partial_{\varepsilon}\, x_{\tau}^{0} \,d B_{\tau}^{k}\right\|^{2} + C\, \mathbb{E} \sup _{t \in[0, T]} \left\|\int_{0}^{t} \eta_{k}\left(x_{\tau}^{0}, \tau\right) d B_{\tau}^{k}\right\|^{2}.
\end{aligned}
$$
For $k=0$, we have
\begin{align*}
& \E \sup_{t \in [0,T]} \left\|\int_{0}^{t} \frac{\partial g_{0}}{\partial x}\left(x_{\tau}^{0}, \tau\right) \partial_{\varepsilon}\, x_{\tau}^{0} d t\right\|^{2} + \E \sup_{t \in [0,T]} \left\|\int_{0}^{t} \eta_{0}\left(x_{\tau}^{0}, \tau\right) d \tau\right\|^{2}, \tag{vi}\\
\leq & \, C \,\E \sup_{t \in [0,T]} \int_{0}^{t}\left\|\frac{\partial g_{0}}{\partial x}\left(x_{\tau}^{0}, t\right)\right\|^{2}\left\|\partial_{\varepsilon}\, x_{\tau}^{0}\right\|^{2} d \tau  +C \E \sup_{t \in [0,T]} \int_{0}^{t}\left\|\eta_{0}\left(x_{\tau}^{0}, \tau\right)\right\|^{2} d \tau, \tag{vii} \\
\leq & \,C \, \E \sup _{s \in[0, T]}\left\|\frac{\partial g_{0}}{\partial x}\left(x_{s}^{0}, s\right)\right\|^{2} \sup_{t \in [0,T]} \int_{0}^{t}\left\|\partial_{\varepsilon}\, x_{\tau}^{0}\right\|^{2} d \tau  +C \, \E \sup_{t \in [0,T]} \int_{0}^{t} C\left(1+\left\|x_{\tau}^{0}\right\|\right)^{2} d \tau, \\
\leq & \,C + C\,\E \sup_{t \in [0,T]} \int_{0}^{t}\left\|\partial_{\varepsilon}\, x_{\tau}^{0}\right\|^{2} d \tau+C \,\E \sup_{t \in [0,T]} \int_{0}^{t}\left\| x_{\tau}^{0}\right\|^{2} d \tau, \tag{viii}\\
\leq & \,C + C\,\int_{0}^{T} \E \sup_{s \in [0,\tau]} \left\|\partial_{\varepsilon}\, x_{s}^{0}\right\|^{2} d \tau+C \, \E \sup_{t \in [0,T]} \left\| x_{t}^{0}\right\|^{2},
\end{align*}
where to get to (vi), we used Jensen's inequality; for (vii), we used the linear growth assumption an $\eta_{0}$, then we obtain (viii) by as derivatives of function $g_0$ are bounded by assumption.\\
Similarly, for $k = 1, \,..., \,m$,
\begin{align*}
& C\, \mathbb{E} \sup _{t \in[0, T]} \left\|\int_{0}^{t} \frac{\partial g_{1}}{\partial x^{i}}\left(x_{\tau}^{0}, \tau\right) \partial_{\varepsilon}\, x_{\tau}^{0} d B_{\tau}\right\|^{2}+ C\, \mathbb{E}\sup _{t \in[0, T]} \left\|\int_{0}^{t} \eta_{1}\left(x_{\tau}^{0}, \tau\right) d B_{\tau}\right\|^{2}, \\
\leq &\, C\,\E \int_{0}^{T}\left\|\frac{\partial g_{1}}{\partial x}\left(x_{\tau}^{0}, \tau\right)\right\|^{2}\left\|\partial_{\varepsilon}\, x_{\tau}^{0}\right\|^{2} d \tau +C\, \mathbb{E} \int_{0}^{T}\left\|\eta_{1}\left(x_{\tau}^{0}, \tau\right)\right\|^{2} d \tau, \tag{ix}\\
\leq &\,  C + C\int^T_0 \E \sup_{s\in[0,\tau]}||\partial_\varepsilon\,x^0_s||^2d\tau + C\, \E \sup_{t\in[0,T]}||x^0_t||^2 \tag{x},
\end{align*}
where we obtain (ix) by the Burkholder-Davis-Gundy inequality and (x) by following similar steps as have shown in (vii) and (viii).\\
We are now ready to sum up each term to acquire a new inequality:
\begin{align*}
\mathbb{E} \sup_{t\in[0,T]} \left\|\partial_{\varepsilon}\, x_{t}^{0}\right\|^{2}  \leq & \,C+C\,  \mathbb{E} \sup _{t \in [0,T]}\left\|x_{t}^{0}\right\|^{2}+C \, \int_{0}^{T} \mathbb{E} \sup _{s \in[0, \tau]}\left\|\partial_{\varepsilon}\, x_{s}^{0}\right\|^{2} d \tau .
\end{align*}

By Gronwall's lemma, we have that
\begin{equation*}
\mathbb{E} \sup _{t \in[0, T]}\left\|\partial_{\varepsilon} \,x_{t}^{0}\right\|^{2} \leq\left(C +C \, \mathbb{E} \sup _{t \in[0, T]}\left\|x^{0}_t\right\|^{2}\right) \exp (C).
\end{equation*}

As it is previously shown that $\mathbb{E} \sup _{t \in[0, \tau]}\left\|x^{\circ}(t)\right\|^{2}$ is bounded, it is clear that $\mathbb{E} \sup _{t \in[0, T]}\left\|\partial_{\varepsilon} \,x_{t}^{0}\right\|^{2}$ is bounded too. 

(3) From similar steps, one can also show that $\mathbb{E} \supp{t \in[0, T]}\left\|\partial_{\varepsilon}^{2}\, x_{t}^{0}\right\|^{2}$ is bounded.
\end{proof}

\textit{Step 3}: Having shown that $x_{t}^{0}, \partial_{\varepsilon}\, x_{t}^{0}, \partial_{\varepsilon}^{2}\, x_{t}^{0}$ are bounded, we proceed to bound the remainder term by proving the following lemma.

\begin{lemma} \label{lem-bd-r}
For a given $\varepsilon \in \mathbb{R}$, let $$\mathcal{R}^{\varepsilon}:= (t, \omega) \mapsto \frac{1}{\varepsilon^{3}}\left(x^{\varepsilon}(t, \omega) - x^{0}(t, \omega) -\varepsilon \partial_{\varepsilon} x^{0} (t, \omega) -\varepsilon^{2} \partial_{\varepsilon}^{2}\, x^{0}(t, \omega)\right),$$ where the stochastic process $x_{t}^{\varepsilon}$ is the solution to Equation (\ref{origSDE}). Then it holds true that

$$
\mathbb{E} \sup _{t \in[0, T]}\left\|\mathcal{R}^{\varepsilon}(t)\right\|^{2} \text { is bounded. }
$$
\end{lemma}

\begin{proof}
The main strategy of this proof is to first rewrite $\varepsilon^{3} \mathcal{R}^{\varepsilon}$ as the sum of some simpler terms and then to bound each term. 
To simplify the notation, we denote $\tilde{x}_{t}^{\varepsilon}$ as $x_{t}^{0}+\varepsilon \partial_{\varepsilon}\,x_{t}^{0}+\varepsilon^{2}\, \partial_{\varepsilon}^{2} x_{t}^{0}$.  \\
For $k=0, .., n$, we define the following terms:
$$
\begin{aligned}
& \theta_{k}(t):=\int_{0}^{t} g_{k}\left(x_{\tau}^{\varepsilon}, \tau\right)-g_{k}\left(\tilde{x}_{\tau}^{\varepsilon}, \tau\right) d B_{\tau}^{k} , \\
& \varphi_{k}(t):=\int_{0}^{t} g_{k}\left(\tilde{x}_{\tau}^{\varepsilon}, \tau\right)-g_{k}\left(x_{\tau}^{0}, \tau\right) -\varepsilon \frac{\partial g_{k}}{\partial x}\left(x_{\tau}^{0}, \tau\right) \partial_{\varepsilon}\, x_{\tau}^{0} -\varepsilon^{2} \Psi_{k}\left(\partial_{\varepsilon}\, x_{\tau}^{0}, x_{\tau}^{0}, \tau\right) -\varepsilon^{2} \frac{\partial g_{k}}{\partial x^{i}}\left(x_{\tau}^{0}, \tau\right) \partial_{\varepsilon}^{2} \,x_{\tau}^{0} d B_{\tau}^{k},\\
& \sigma_{k}(t):= -\varepsilon \int_{0}^{t}\eta_{k}\left(x_{\tau}^{0}, \tau\right)+2 \varepsilon \, \frac{\partial \eta}{\partial x}\left(x_{\tau}^{0}, \tau\right) \partial_{\varepsilon}\, x_{\tau}^{0} d B_{\tau}^{k}.
\end{aligned}
$$

Hence, we have $\varepsilon^{3} \mathcal{R}^{\varepsilon}(t)=\sum_{k=0}^{1} \theta_{k}(t)+\varphi_{k}(t)+\sigma_{k}(t)$.

For $\theta_k(t)$, we have
\begin{align*}
\mathbb{E} \sup_{t \in[0, T]} \left\|\theta_{k}(t)\right\|^{2} & \leq C\, \mathbb{E} \sup _{t \in[0, T]} \int_{0}^{t}\left\|g_{k}\left(x_{\varphi}^{\varepsilon}, e\right)-g_{k}\left(\tilde{x}_{\varphi}^{\varepsilon}, \tau\right)\right\|^{2} d \tau, \tag{i}\\
& \leq C \, \int_{0}^{T} \mathbb{E} \sup _{t \in[0, tau]} \left\|x_{t}^{\varepsilon}-\tilde{x}_{t}^{\varepsilon}\right\|^{2} d \tau, \tag{ii}\\
& \leq C \, \int_{0}^{T} \mathbb{E} \sup _{t \in[0, \tau]}\left\|\mathcal{R}^{\varepsilon}(t)\right\|^{2} d \tau, \tag{iii},
\end{align*}
where to obtain (i) we used Jensen's inequality when $k = 0$ and by the Burkholder-Davis-Gundy inequality when $k=1$, used the Lipschitz condition of $g_{k}$ to obtain (ii), and for (iii), it is because $\varepsilon^{3} \mathcal{R}^{\varepsilon}(t)=\tilde{x}_{t}^{\varepsilon}-x_{t}^{\varepsilon}.$\\
We note that from Taylor's theorem, for any $s \in[0, t], \, k=0,1$, there exists some $\varepsilon_{s} \in(0, \varepsilon)$ s.t.
\begin{equation}\label{taylor}
g_{k}\left(\tilde{x}_{s}^{\varepsilon}, s\right)-g_{k}\left(x_{s}^{0}, s\right)-\varepsilon \frac{\partial g_{k}}{\partial x}\left(x_{s}^{0}, s\right) \partial_{\varepsilon} x_{s}^{0}  =\varepsilon^{2} \frac{\partial g_{k}}{\partial x}\left(\tilde{x}_{s}^{\varepsilon_{s}}\right) \partial_{\varepsilon}^{2}\, x_{s}^{0}+\varepsilon^{2} \Psi\left(\partial_{\varepsilon}\, x_{s}^{0}, \tilde{x}_{s}^{\varepsilon_{s}}, s\right).
\end{equation}

For $\varphi_{k}(t)$, we have 

\begin{align*}
& \mathbb{E} \sup _{t \in[0, T]}\left\|\varphi_{k}(t)\right\|^{2} \\
\leq \, & C\, \mathbb{E} \sup _{t \in[0, T]} \int_{0}^{t} \| \frac{\partial g_{k}}{\partial x}\left(\tilde{x}_{s}^{\varepsilon_{s}}\right) \partial_{\varepsilon}^{2} \,x_{s}^{0}+\Psi_{k}\left(\partial_{\varepsilon}\, x_{s}^{0}, \tilde{x}_{s}^{\varepsilon_{s}}, s\right) -\frac{\partial g_{k}}{\partial x}\left(x_{s}^{0}\right) \partial_{\varepsilon}^{2}\, x_{s}^{0}-\Psi_{k}\left(\partial_{\varepsilon}\, x_{s}^{0}, x_{s}^{0}, s\right) \|^{2} d s, \tag{iv}\\
\leq \, & \,  C\, \mathbb{E} \sup _{t \in[0, T]} \int_{0}^{t} \left\| \frac{\partial g_{k}}{\partial x}\left(\tilde{x}_{s}^{\varepsilon_{s}}\right) -  \frac{\partial g_{k}}{\partial x}\left(x_{s}^{0}\right)  \right\|^2 \left\|\partial_{\varepsilon}^{2} \,x_{s}^{0} \right\|^2
 +\left\|\Psi_{k}\left(\partial_{\varepsilon} x_{s}^{0}, \tilde{x}_{s}, s\right)-\Psi_{k}\left(\partial_{\varepsilon} x_{s}^{0}, x_{s}^{0}, s\right)\right\|^{2} d s, \tag{v}\\
\leq \, & C\, \E \sup _{t \in[0, T]} \int_{0}^{t}\left\|\tilde{x}_{s}^{\varepsilon_{s}}-x_{s}^{0}\right\|^{2}\left(C+\left\|\partial_{\varepsilon}^{2}\, x_{s}^{0}\right\|^{2}\right) d s,\tag{vi}\\
\leq \, & C\, \E \sup _{t \in[0, T]} \int_{0}^{t}\left\|\varepsilon \partial_{\varepsilon}\, x^{0}_s + \varepsilon^2 \partial^2_{\varepsilon} \, x^{0}_s \right\|^{2}\left(C+\left\|\partial_{\varepsilon}^{2}\, x_{s}^{0}\right\|^{2}\right) d s,\\
\leq \, & C \left(\E \sup _{t \in[0,T]} \left\| \partial_{\varepsilon}\, x_{s}^{0}\right\|^{2}) + \E \sup _{t \in[0,T]} \left\| \partial_{\varepsilon}^{2}\, x_{s}^{0}\right\|^{2})\right)\left(C+\E \sup _{t \in[0,T]} \left\| \partial_{\varepsilon}^{2}\, x_{s}^{0}\right\|^{2}\right)\tag{vii},
\end{align*}
where for (iv), we used Equation (\ref{taylor}) and Jensen's  inequality for $k=0$ and the Burkholder-Davis-Gundy inequality for $k=1$; to obtain (v), we applied Jensen's equality; we then derived (vi) from the Lipschitz conditions of $g_{k}$ and $\Psi_{k}$; and finally another application of Jensen's inequality gives (vii) which is bounded as a result from the Lemma \ref{lem-bd}.\\

For $\sigma_{k}(t)$,
\begin{align*}
\sup _{t \in[0, T]}\left\|\sigma_{0}(t)\right\|^{2} \leq\, & C \, \varepsilon \, \int_{0}^{T} \mathbb{E} \sup _{s \in[0, t]}\left\|\eta_{k}\left(x_{s}^{0}, s\right)\right\|^{2}+C \mathbb{E} \sup _{s \in[0, t]}\left\|\frac{\partial \eta_{k}}{\partial x}\left(x_{s}^{0}, s\right)\right\|^{2}\left\|\partial_{\varepsilon} \,x_{s}^{0}\right\|^{2} d t, \tag{ix}\\
\leq\, & C\, \int_0^T C \left( 1+ \E \sup _{s \in [0, t]} \left\| x_s^0\right\|^2 \right)+ C \mathbb{E} \sup _{t \in[0, T]} \left\|\frac{\partial \eta_{k}}{\partial x}\left(x_{t}^{0}, t\right)\right\|^{2} \int_{0}^{T} \mathbb{E} \sup _{s \in[0, t]}\left\|\partial_{\varepsilon} x_{s}^{0}\right\|^{2} d t, \tag{x}\\
\leq \, & c +C \, \mathbb{E} \sup _t\in[0,T]\left\|x_{s}^{0}\right\|^{2} + C \,\mathbb{E} \sup _{t \in [0, T]}\left\|\frac{\partial \eta}{\partial x}\left(x_{t}^{0}, t\right)\right\|^{2} \mathbb{E} \sup _{t \in [0, T]}\left\|\partial_{\varepsilon} x_{t}^{0}\right\|^{2},\tag{xi}
\end{align*}
where we obtained (ix) by Jensen's inequality when $k=0$ and by Burkholder-Davis-Gundy inequality when $k=1$, and (x) by the linear growth assumption on $\eta_{k}$; one can see that (xi) is bounded by recalling the Lemma \ref{lem-bd} and the assumption that $\eta_{k}$ has bounded derivatives.

Hence, by Jensen's inequality and Gronwall's lemma, we have 
\begin{align*}
 \mathbb{E} \sup _{t \in[0, T]}\left\|\mathcal{R}^{\varepsilon}(t)\right\|^{2} \leq\, & C\, \sum_{k=0}^{K}\, \mathbb{E} \sup _{t \in[0, T]}\left\|\theta_{k}(t)\right\|^{2} +\mathbb{E} \sup _{t \in[0, T]}\left\|\varphi_{k}(t)\right\|^{2}  +\mathbb{E} \sup _{t \in [0, T]}\left\|\sigma_{k}(t)\right\|^{2}, \\
\leq \, & C +C \, \int_{0}^{T} \mathbb{E} \sup _{t \in[0, \tau]}\left\|\mathcal{R}^{\varepsilon}(t)\right\|^{2} d \tau, \\
\leq \, & C \exp \left(C \right).
\end{align*}
Therefore, $\mathbb{E} \sup \left\|\mathcal{R}^{\varepsilon}(t)\right\|^{2}$ is bounded.

\end{proof}

Finally, it is now straightforward to show Equation (\ref{taylor-1}) by applying a second-order Taylor expansion on $f\left(x_{t}^{0}+\varepsilon \partial_{\varepsilon} x_{t}^{0}+\varepsilon^{2} \partial_{\varepsilon}^{2} x_{t}^{0}\right.$ $\left.+\varepsilon^{3} R^{\varepsilon}(t)\right)$.

\end{proof}

We are now ready to show Theorem \ref{imp-reg-thm}. One notes that Corollary \ref{unstable-imp-thm} directly follows from the result too.
\begin{proof} (\textit{Theorem \ref{imp-reg-thm}})
From Proposition \ref{main}, it is noteworthy to point out that the derived SDEs (\ref{prop-SDEs}) for $\partial_{\varepsilon}\, x^0_t$ and $\partial_{\varepsilon}^2\, x_t^0$ are vector-valued general linear SDEs. With some steps of derivations, one can express the solutions as: 
\begin{align*}
\partial_{\varepsilon}\, x^0_t =  \, & \Phi_t \int_0^t \Phi_s^{-1} \left( \eta_0(x_s^0, s) - \sum _{k=1}^m \frac{\partial g_k}{\partial x}(x_s^0, s)\eta_k(x_s^0, s)\right)ds + \, \Phi_t \int_0^t \Phi_s^{-1} \eta_k(x_s^0, s) dB_s^k\tag{a}\label{sol-a}\\
\partial_{\varepsilon}^2\, x_t^0  =\, & \Phi_t \int_0^t \Phi_s^{-1} \biggl( \Psi_0(x_s^0, \partial_\varepsilon\, x_s^0, s) + 2\,\frac{\partial \eta_0}{\partial x}(x_s^0, s) \partial_{\varepsilon}\,x_s^0  \\ & \,\,\,\,\,\,\,\,\,- \sum _{k=1}^{m} \frac{\partial g_k}{\partial x}(x_s^0, s) \Bigl( \Psi_k(x_s^0, \partial_\varepsilon\, x_s^0, s) + 2\,\frac{\partial \eta_k}{\partial x}(x_s^0, s) \partial_{\varepsilon}\,x_s^0) \Bigr) \biggr) ds,\\
\,& + \Phi_t \int_0^t \Phi_s^{-1} \sum _{k=1}^{m} \left( \Psi_k(x_s^0, \partial_\varepsilon\, x_s^0, s) + 2\,\frac{\partial \eta_k}{\partial x}(x_s^0, s) \partial_{\varepsilon}\,x_s^0 \right)dB_s^k\tag{b}\label{sol-b},
\end{align*}
where $n\times n$ matrix $\Phi_t$ is the fundamental matrix of the corresponding homogeneous equation:
\begin{equation}
    d \Phi_t = \frac{\partial g_k}{\partial x}(x_t^0, t)\,\Phi_t\, dB^k_t,
\end{equation}
with initial value 
\begin{equation}
    \Phi(0) = I.
\end{equation}
It is worthy to note that the fundamental matrix $\Phi_t$ is non-deterministic and when $\frac{\partial g_i}{\partial x}$ and $\frac{\partial g_j}{\partial x}$ commutes, $\Phi_t$ has explicit solution 
\begin{equation}
    \Phi_t = \exp \left( \int_0^t \frac{\partial g_k}{\partial x} (x_s^0, s)dB_s^k - \frac{1}{2}\int_0^t \frac{\partial g_k}{\partial x}(x_s^0, s)\frac{\partial g_k}{\partial x}(x_s^0, s)^\top ds \right).
\end{equation}

Having obtained the explicit solutions, one can plug in corresponding terms and obtain the results of \textit{Theorem \ref{imp-reg-thm}}) after a Taylor expansion of the loss function $\mathcal{L}$.
\end{proof}

\newpage
\section{Error Accumulation During the Inference Phase and its Effects to Value Functions \label{error-pred}}
\begin{theorem}  (Error accumulation due to initial representation error )\\\label{error-accu}
Let $\delta := \E \,\|\varepsilon\|$ and $d_{\varepsilon} := \mathbb{E} \sup _{t \in [0, T]} \left\|h_{t}^{\varepsilon}-h_{t}^{0}\right\|^2 + \left\|\tilde{z}_{t}^{\varepsilon}-\tilde{z}_{t}^{0}\right\|^2$. It holds that as $\delta \rightarrow 0$,
\begin{equation}
d_{\varepsilon} \leq   \, \delta\, C \left( \mathcal{J}_0  + \mathcal{J}_1 \right) +  \, \delta^2 \,C \left(\exp \left( \, \mathcal{H}_0  \left( \mathcal{J}_0  + \mathcal{J}_1 \right) \right)  + \exp \left( \, \mathcal{H}_1  \left( \mathcal{J}_0  + \mathcal{J}_1 \right) \right) \right)+ \mathcal{O}(\delta^3),
\end{equation} 
where
\begin{align*}
    \mathcal{J}_0 = & \exp \left( \mathcal{F}_h + \mathcal{F}_z + \mathcal{P}_h \right), \,
    \mathcal{J}_1 = \exp \left( \bar{\mathcal{P}}_h \right),\\
    \mathcal{H}_0 = & \mathcal{F}_{hh} + \mathcal{F}_{hz} + \mathcal{F}_{zh}+ \mathcal{F}_{zz} +  \mathcal{P}_{hh},\,
    \mathcal{H}_1 = \bar{\mathcal{P}}_{hh} 
\end{align*}
\begin{align*}
    \mathcal{F}_h =&  C \,\E \sup_{t \in [0,T]} \left\| \frac{\partial f}{\partial h} +  \frac{\partial f}{\partial a} \partial_h\rho  \right\|_F^2,\,\,\,\,
     \mathcal{F}_z =  C \, \E \sup_{t \in [0,T]} \left\| \frac{\partial f}{\partial z} +  \frac{\partial f}{\partial a} \partial_z\rho \right\|_F^2,\,\\
    \mathcal{P}_h = & C \, \E \sup_{t \in [0,T]} \left\| \frac{\partial p}{\partial h}\right\|_F^2 , \,
     \bar{\mathcal{P}}_h =  C \, \E \sup_{t \in [0,T]} \left\| \frac{\partial \bar{p}}{\partial h}\right\|_F^2 , \\
     \mathcal{F}_{hh} = & C\, \E \sup_{t \in [0,T]} \left\| \frac{\partial^2 f}{\partial h^2} + \frac{\partial^2 f}{\partial h\partial a} \partial_h\rho + \frac{\partial f}{\partial a} \partial^2_{hh}\rho \right\|^2_F,\\
     \mathcal{F}_{hz} = &C\, \E \sup_{t \in [0,T]} \left\| \frac{\partial^2 f}{\partial h \partial z} + \frac{\partial^2 f}{\partial z\partial a} \partial_h\rho + \frac{\partial f}{\partial a} \partial^2_{zh}\rho\right\|^2_F \\
      \mathcal{F}_{zh} = &C\, \E \sup_{t \in [0,T]} \left\| \frac{\partial^2 f}{\partial h \partial z} + \frac{\partial^2 f}{\partial h\partial a} \partial_z\rho + \frac{\partial f}{\partial a} \partial^2_{hz}\rho\right\|^2_F \\
     \mathcal{F}_{zz} = & C\, \E \sup_{t \in [0,T]} \left\| \frac{\partial^2 f}{\partial z^2} + \frac{\partial^2 f}{\partial z\partial a} \partial_z\rho + \frac{\partial f}{\partial a} \partial^2_{zz}\rho\right\|^2_F, \\
     \mathcal{P}_{hh} = &C\, \E \sup_{t \in [0,T]} \left\| \frac{\partial^2 p}{\partial h^2}\right\|^2_F, \,\, \bar{\mathcal{P}}_{hh} = C\, \E \sup_{t \in [0,T]} \left\| \frac{\partial^2 \bar{p}}{\partial h^2} \right\|^2_F,
\end{align*}

where for brevity, when functions always have inputs $(\tilde{z}_t^0, h_t^0, t)$, we adopt the shorthand to write, for example, $f(\tilde{z}_t^0, h_t^0, t)$ as $f$.
\end{theorem}

Before proving the main result \ref{error-accu}, we first show the general case of perturbation in initial values.  Consider the following general system with noise at the initial value:
\begin{align}\label{origSDE-1}
& d x_{t}=g_{0}\left(x_{t}, t\right) d t+ g_{k}\left(x_{t}, t\right) d B_{t}^{k} , \\
& x(0)=x_{0}+\varepsilon,
\end{align}
where the initial perturbation $\varepsilon \in \mathbb{R}^{n} \times \Omega$. As $g_{k}$ are $\mathcal{C}_{g}^{2, \alpha}$ functions, by the classical result on the existence and the uniqueness of solution to SDE, there exists a unique solution to Equation (\ref{origSDE-1}), denoted as $x_{t}^{\varepsilon}$ or $x^{\varepsilon}(t)$.

To simplify the notation, we write $\partial_{i}\, x_{t}^{\varepsilon}:=\frac{\partial x^{\varepsilon}(t)}{\partial x^{i}}, \partial_{i j}^{2}\, x_{t}^{\varepsilon}=\frac{\partial^{2} x_{t}^{\varepsilon}}{\partial x^{i} \partial x^{j}}$, for $i, j=1, \,\ldots, \,n$ that are, respectively, the first and second-order derivatives of the solution $x^{\varepsilon}(t)$ w.r.t. the changes in the corresponding coordinates of the initial value.
When $\varepsilon=0 \in \mathbb{R}^{n}$, we denote the solutions to Equation (\ref{origSDE-1}) as $x_{t}^{0}$ with its first and second derivatives $\partial_{i}\, x_{t}^{0}, \partial_{i j}^{2}\, x_{t}^{0}$, respectively.

\begin{proposition}

\label{main-2}
Let $\delta := \E \,\|\varepsilon\|$, it holds that
\begin{align}
\mathbb{E} \sup _{t \in [0, T]} \left\|x_{t}^{\varepsilon}-x_{t}^{0} \right\|^2
 \leq  \sum_{k=0,1} C\, \delta \left( C\, \E \sup_{t \in [0,T]} \left\| \frac{\partial g_k}{\partial x} (x_t^0, t)\right\|_F^2 \right) \nonumber \\  +  C\, \delta^2 \exp \left( C\, \E \sup_{t \in [0,T]} \left\| \frac{\partial^2 g_k}{\partial x^2}(x^0_t, t) \right\|_F^2 \sum_{\bar{k}=0,1} \exp \left( C \,\E \sup_{t \in [0,T]} \left\| \frac{\partial g_{\bar{k}}}{\partial x} (x_t^0, t)\right\|_F^2 \right)\right) + \mathcal{O}(\delta^3), 
\end{align}
as $\delta \rightarrow 0.$

\end{proposition}

\begin{proof}
Similar to the previous section, for notational convenience, we write $t$ as $B_{t}^{0}$ and employs Einstein summation notation. Hence, Equation  (\ref{origSDE-1}) can be shorten as
\begin{equation}\label{sde-1-short}
d x_{t}=g_{k}\left(x_{t}, t\right) d B_{t}^{k},
\end{equation}
with initial values $x(0) = x_{0}+\varepsilon$.

To begin, we find the SDEs that characterize $\partial_{i}\, x_{t}^{\varepsilon}$ and $\partial_{i j}^{2}\, x_{t}^{\varepsilon}$, for $i, \,j=1,\,..., \,n$.

For $\partial_{i}\, x^{\varepsilon}_t$, we apply Theorem 3.1 from Section 2 in \cite{Hennequin_Dudley_Kunita_Ledrappier_1984} on Equation (\ref{sde-1-short}) and $\partial_{i}\, x_{t}^{\varepsilon}$ satisfy the following SDE with probability 1,

\begin{equation}\label{sde-1st-r}
d \partial_{i}\, x_{t}^{\varepsilon}=\frac{\partial g_{k}}{\partial x}\left(x_{t}^{\varepsilon}, t\right) \partial_{i} \,x_{t}^{\varepsilon} d B_{t}^{k}
\end{equation}

with initial value $\partial_{i} x^{\varepsilon}_0$ to be the unit vector $e_{i}=$ $(0,\,0, \,\ldots, \,1, \,\ldots, \,0)$ that is all zeros except one in the $i^{\text{th}}$ coordinate.

For $\partial_{i j}^{2}\, x^{\varepsilon}_t$, we again apply Theorem 3.1 from Section 2 in \cite{Hennequin_Dudley_Kunita_Ledrappier_1984}  on the SDE (\ref{sde-1st-r}) and obtain that $\partial_{i j}^{2} x_{b}^{\varepsilon}$ satisfy the following SDE with probability 1,
\begin{equation}\label{sde-2nd-r}
d \partial_{i j}^{2}\, x_{t}^{\varepsilon}=\Psi_{k}\left(x_{t}^{\varepsilon}, \partial_{i}\, x_{t}^{\varepsilon}, t\right) \partial_{i j}^{2}\, x_{t}^{\varepsilon} d B_{t}^{k},
\end{equation}
with the initial value $\partial_{i j}\,x^{\varepsilon}(0)=e_{j}$, where 
\begin{equation*}
\Psi_{k}: \mathbb{R}^{d} \times \mathbb{R}^{d} \times[0, T] \rightarrow \mathbb{R}^{d \times d},\,
(x, \partial_i\,x, t) \mapsto\left(\frac{\partial^{2} g_{k}^{l}}{\partial x^{u} \partial x^{v}}\left(x^{\varepsilon}_t, t\right)\right)_{l, u, v} \partial_i\,x^{v}.
\end{equation*}

For the next step, we show that with probability 1, the following holds
\begin{equation}\label{some-taylor}
x_{t}^{\varepsilon}=x^{0}_t+\varepsilon^i \,\partial_i\, x_{t}^{0} +\frac{1}{2} \, \varepsilon^{i} \varepsilon^{j} \,\partial_{i j}^{2}\, x_{t}^{0}+O\left(\varepsilon^{3}\right),
\end{equation}
as $\left\|\varepsilon\right\| \rightarrow 0.$ \\
 One can follow the similar steps of proofs for Lemma (\ref{lem-bd}) and (\ref{lem-bd-r}) in the previous section to show that $\mathbb{E} \sup _{t \in[0, T]}\left\|x_{t}^{0}\right\|^{2},\, \mathbb{E} \sup _{t \in[0, T]}\left\|\partial_{i} x_{t}^{0}\right\|^{2}, \,\mathbb{E} \sup _{t \in[0, T]}\left\|\partial_{i j}^{2} x_{t}^{0}\right\|^{2}\,$ and the remainder term are bounded. Hence, Equation (\ref{some-taylor}) holds with probability 1.\\

Indeed, for $\E \sup_{t \in [0,T]} \left\| \partial_i\, x_t^0\right\|^2$, it holds that
\begin{align}
\E \sup_{t \in [0,T]} \left\| \partial_i\, x_t^0\right\|^2 \leq \, & C \left\| e_i \right\|^2 + \sum_{k=0, 1} \E \sup_{t \in [0, T]} C \int_0^t \left\| \frac{\partial g_k}{\partial x}(x_s^0, s)\right\|_F^2\left\| \partial_i\,x_s\right\|^2ds\\ 
\leq \, & \sum_{k=0, 1} C \exp \left( C\, \E \sup_{t \in [0,T]} \left\| \frac{\partial g_k}{\partial x} (x_t^0, t)\right\|_F^2 \right). \label{ineq-1}
\end{align}

Similarly, for $\E \sup_{t \in [0,T]} \left\| \partial^2_{ij}\, x_t^0\right\|^2$, it holds that
\begin{align}\label{ineq-2}
\E \sup_{t \in [0,T]} \left\| \partial^2_{ij}\, x_t^0\right\|^2 \leq \,&
C \left\| e_i \right\|^2 + \sum_{k=0, 1} \E \sup_{t \in [0, T]} C \int_0^t \left\| \frac{\partial^2 g_k}{\partial x^2}(x^0_s, s)\right\|_F^2 \left\| \partial_i \,x_s^0\right\|^2 \left\| \partial_{ij}^2\, x_s^0\right\|^2ds\\
\leq \, & C \sum_{k=0}^{1} \exp \left( C\, \E \sup_{t \in [0,T]} \left\| \frac{\partial^2 g_k}{\partial x^2}(x^0_t, t) \right\|_F^2 \left\|\partial_i\,x_t^0\right\|^2 \right)\\
\leq \, & C \sum_{k=0, 1} \exp \left( C\, \E \sup_{t \in [0,T]} \left\| \frac{\partial^2 g_k}{\partial x^2}(x^0_t, t) \right\|_F^2\exp \left( C \,\E \sup_{t \in [0,T]} \left\| \frac{\partial g_k}{\partial x} (x_t^0, t)\right\|_F^2 \right)\right).
\end{align}

Therefore, we could obtain the proposition by applying Jensen's inequality to Equation (\ref{some-taylor}) and plugging with \ref{ineq-1} and \ref{ineq-2}.
\end{proof}

Now we are ready to prove Theorem \ref{error-accu}. We note that one could then obtain Corollary \ref{acc-val-thm} without much more effort by a standard application of Taylor's theorem. 
\begin{proof} (Proof for Theorem \ref{error-accu})

At $(h_t, \tilde{z}_t, \pi(h_t, \tilde{z}_t))$,  where the local optimal policy $\pi(h_t, \tilde{z}_t)$, denoted as $a_t^*$, there exists an open neighborhood $V \subseteq \mathcal{A}$  of $a^*_t$ such that $a_t^*$ is the local maximizer for $Q(h_t, \tilde{z}_t, \cdot)$ by definition. Then, $\pf{Q}{a}(h_t, \tilde{z}_t, a^*_t) = 0$,  and $\pft{Q}{a}(h_t, \tilde{z}_t, a)$ is negative definite. As $\pft{Q}{a}$ is non-degenerate in the neighborhood $V$, by the implicit function theorem,  there exists a neighborhood $U\times V$ of $(h_t, \tilde{z}_t, a^*_t)$ such that there exists a $\mathcal{C}^2$ map  $\rho: U \rightarrow V$ such that $\pf{Q}{a}(h, \tilde{z}, \rho(h, \tilde{z})) = 0$  and $\rho(h, \tilde{z})$ is the local maximizer of $Q(h, \tilde{z}, \cdot)$ for any $h, \tilde{z} \in U$. Furthermore, we have that $\partial_h\,{\rho}  = -\pft{Q}{a}^{-1}\pf{^2Q}{a\partial h}.$
Similarly, other first-terms and second-order terms $\partial_z \rho$, $\partial^2_{zz} \rho, \partial^2_{zh} \rho, \partial^2_{hz} \rho, \partial^2_{hh} \rho$ can be explicitly expressed without much additional effort (e.g., in \cite{Loomis2014-ux}, \cite{Cartan2017-mw}). 

The rest of the proof is easy to see after plugging in the corresponding terms from Proposition \ref{main-2}.
\end{proof}

\newpage
\section{Experimental Details\label{exp-appendix}}
In this section, we provide additional details and results beyond thoese in the main paper.

\subsection{Model Implementation and Training \label{model-appendix}}
Our baseline is based on the DreamerV2 Tensorflow implementation. Our theoretical and empirical results should not matter on the choice of specific version; so we chose DreamerV2 as its codebase implementation is simpler than V3. We incorporated a computationally efficient approximation of the Jacobian norm for the sequence model, as detailed in \cite{hoffman2019robust}, using a single projection. During our experiments, all models were trained using the default hyperparameters (see Table \ref{table:hyperparameters}) for the MuJoCo tasks. The training was conducted on an NVIDIA A100 and a GTX 4090, with each session lasting less than 15 hours.

\begin{table}[h!]
\centering
\begin{tabular}{|l|l|}
\hline
\textbf{Hyperparameter} & \textbf{Value} \\ \hline
eval\_every & 1e4 \\ \hline
prefill & 1000 \\ \hline
train\_every & 5 \\ \hline
rssm.hidden & 200 \\ \hline
rssm.deter & 200 \\ \hline
model\_opt.lr & 3e-4 \\ \hline
actor\_opt.lr & 8e-5 \\ \hline 
replay\_capacity & 2e6 \\ \hline
dataset\_batch & 16 \\ \hline
precision & 16 \\ \hline
clip\_rewards & tanh \\ \hline
expl\_behavior & greedy \\ \hline
encoder\_cnn\_depth & 48 \\ \hline
decoder\_cnn\_depth & 48 \\ \hline
loss\_scales\_kl & 1.0 \\ \hline
discount & 0.99 \\ \hline
jac\_lambda & 0.01 \\ \hline
\end{tabular}
\caption{Hyperparameters for DreamerV2 model.}
\label{table:hyperparameters}
\end{table}
\newpage
\subsection{Additional Results on Generalization on Perturbed States}
In this experiment, we investigated the effectiveness of Jacobian regularization in model trained against a baseline during the inference phase with perturbed state images. We consider three types of perturbations:  (1) Gaussian noise across the full image, denoted as $\mathcal{N}(\mu_1, \sigma_1^2)$ ; (2) rotation; and (3) noise applied to a percentage of the image, $\mathcal{N}(\mu_2, \sigma_2^2).$ (In Walker task, $\mu_1=\mu_2=0.5, \sigma_2^2=0.15$; in Quadruped task, $\mu_1=0, \mu_2=0.05, \sigma_2^2=0.2$.) In each case of perturbations, we examine a collection of noise levels: (1) variance $\sigma^2$ from $0.05$ to $0.55$; (2) rotation degree $\alpha$ $20$ and $30$; and (3) masked image percentage $\beta\%$ from $25$ to $75$. 
\subsection{Walker Task}
\begin{table}[!htb]
    \centering
    \begin{tabular}{|c|c|c|c|c|}
    \toprule
         $\beta \%$ mask, $\mathcal{N}(0.5, 0.15)$&  mean (with Jac.) &  stdev (with Jac.)&  mean (baseline)& stdev (baseline)\\
         \hline
         25\% &  882.78&  28.57199976&  929.778& 10.13141451\\
         30\% &  878.732&  40.92085898&  811.198& 7.663919934\\
         35\%&  856.32&  37.56882045&  799.98& 29.75286097\\
         40\%&  804.206&  47.53578989&  688.382& 43.21310246\\
         45\%&  822.97&  80.36907477&  601.862& 42.49662057\\
         50\% &  725.812&  43.87836335&  583.418& 76.49237076\\
         55\% &  768.68&  50.71423045&  562.574& 59.88315135\\
         60\% &  730.864&  23.37324967&  484.038& 90.38940234\\
         65\% &  696.936&  65.26307708&  516.936& 41.44549462\\
         70\%&  687.346&  70.9078686&  411.922& 45.85808832\\
         75\% &  685.492&  63.22171723&  446.74& 40.66898799\\
\bottomrule
    \end{tabular}
    \caption{\textit{Walker.} Mean and standard deviation of accumulated rewards under masked perturbation of increasing percentage.}
    \label{tab:my_label}
\end{table}

\begin{table}[!htb]
    \centering
    \begin{tabular}{|c|c|c|c|c|}
    \toprule
         full, $\mathcal{N}(0.5, \sigma^2)$&  mean (with Jac.) &  stdev (with Jac.)&  mean (baseline)& stdev (baseline)\\
         \hline
         0.05&  894.594&  39.86907737&  929.778& 40.91\\
         0.10&  922.854&  27.28533819&  811.198& 98.79\\
         0.15&  941.512&  16.47165049&  799.98& 106.01\\
         0.20&  840.706&  66.12470628&  688.382& 70.78\\
         0.25&  811.764&  75.06276427&  601.862& 83.65\\
         0.30 &  779.504&  53.29238107&  583.418& 173.59\\
         0.35 &  807.996&  34.35949621&  562.574& 79.30\\
         0.40 &  751.986&  85.20137722&  484.038& 112.43\\
         0.45 &  663.578&  60.18862658&  516.936& 90.25\\
         0.50&  618.982&  61.10094983&  411.922& 116.94\\
         0.55  &  578.62&  64.25840684&  446.74& 84.44\\
\bottomrule
    \end{tabular}
    \caption{\textit{Walker.} Mean and standard deviation of accumulated rewards under Gaussian perturbation of increasing variance.}
    \label{tab:my_label}
\end{table}

\begin{table}[!htb]
    \centering
    \begin{tabular}{|c|c|c|c|c|}
    \toprule
         rotation, $\alpha^\circ$&  mean (with Jac.) &  stdev (with Jac.)&  mean (baseline)& stdev (baseline)\\
         \hline
         20 &  423.81& 12.90174678&  391.65& 35.33559636\\
         30 &  226.04&  23.00445979&  197.53&15.26706914\\
\bottomrule
    \end{tabular}
    \caption{\textit{Walker.} Mean and standard deviation of accumulated rewards under rotations.}
    \label{tab:my_label}
\end{table}
\newpage
\subsection{Quardruped Task}
\begin{table}[!htb]
    \centering
    \begin{tabular}{|c|c|c|c|c|}
    \toprule
         $\beta \%$ mask, $\mathcal{N}(0.5, 0.15)$&  mean (with Jac.) &  stdev (with Jac.)&  mean (baseline)& stdev (baseline)\\
         \hline
         25\% &  393.242&  41.10002579&  361.764& 81.41175179\\
         30\% &  384.11&  20.70463958&  333.364& 101.7413185\\
         35\%&  354.222&  53.14855379&  306.972& 16.02275164\\
         40\%&  329.404&  39.1193856&  266.088& 51.20298351\\
         45\%&  360.662&  36.86801622&  281.342& 47.85950867\\
         50\% &  321.556&  27.66758085&  222.222& 22.0668251\\
         55\% &  300.258&  31.44931987&  203.578& 14.38754218\\
         60\% &  321&  18.42956321&  217.98& 23.81819368\\
         65\% &  304.62&  20.75493676&  209.238& 47.14895407\\
         70\%&  301.166&  18.2485583&  193.514& 60.83781004\\
         75\% &  304.92&  18.63214963&  169.58& 30.83637462\\
\bottomrule
    \end{tabular}
    \caption{\textit{Quadruped.} Mean and standard deviation of accumulated rewards under masked perturbation of increasing percentage.}
    \label{tab:my_label}
\end{table}

\begin{table}[!htb]
    \centering
    \begin{tabular}{|c|c|c|c|c|}
    \toprule
         full, $\mathcal{N}(0, \sigma^2)$&  mean (with Jac.) &  stdev (with Jac.)&  mean (baseline)& stdev (baseline)\\
         \hline
         0.10&  416.258&  20.87925573&  326.74& 40.30425536\\
         0.15&  308.218&  24.26432093&  214.718& 15.7782198\\
         0.20&  314.29&  44.73612075&  218.756& 35.41520832\\
         0.25&  293.02&  24.29582269&  190.78& 26.22250465\\
         0.30 &  269.778&  21.83423047&  207.336& 39.1071161\\
         0.35 &  282.046&  13.55303767&  217.048& 29.89589972\\
         0.40 &  273.814&  19.81361476&  190.208& 59.61166975\\
         0.45 &  267.18&  17.5276068&  195.606& 18.91137964\\
         0.50&  268.838&  29.45000543&  194.082& 26.76677642\\
         0.55  &  252.54&  22.516283&  150.786& 24.53362855\\
\bottomrule
    \end{tabular}
    \caption{\textit{Quadruped.} Mean and standard deviation of accumulated rewards under Gaussian perturbation of increasing variance.}
    \label{tab:my_label}
\end{table}

\begin{table}[!htb]
    \centering
    \begin{tabular}{|c|c|c|c|c|}
    \toprule
         rotation, $\alpha^\circ$&  mean (with Jac.) &  stdev (with Jac.)&  mean (baseline)& stdev (baseline)\\
         \hline
         20 &  787.634&  101.5974723&  681.032& 133.7507948\\
         30 &  610.526&  97.74499159&  389.406& 61.5997198\\
\bottomrule
    \end{tabular}
    \caption{\textit{Quadruped.} Mean and standard deviation of accumulated rewards under rotations.}
    \label{tab:my_label}
\end{table}

\newpage
\subsection{Additional Results on Robustness against Encoder Errors}
In this experiment, we evaluate the robustness of model trained with Jacobian regularization against two exogenous error signals (1) zero-drift error with  $\mu_t = 0, \sigma_t^2$ ($\sigma_t^2=5$ in Walker, $\sigma_t^2=0.1$ in Quadruped), and (2) non-zero-drift error with $\mu_t \sim [0, 5], \sigma_t^2 \sim [0,5]$ uniformly. $\lambda$ weight of Jacobian regularization is $0.01$. In this section, we included plot results of both evaluation and training scores. 
\subsubsection{Walker Task}
\begin{figure}[!htb]
    \centering
    \includegraphics[width=1\linewidth]{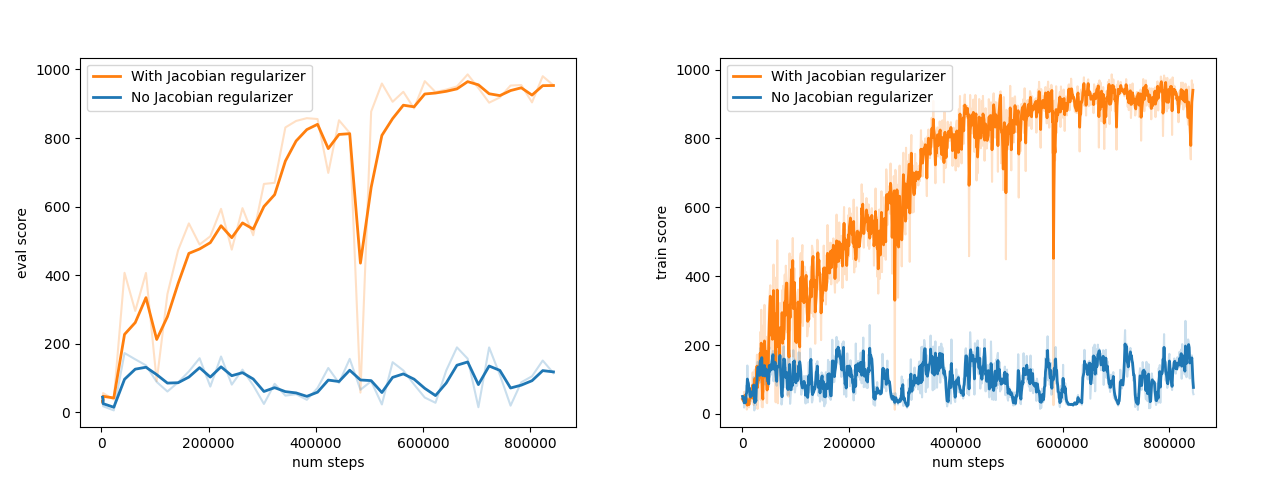}
    \caption{\textit{Walker.} Eval (left) and train scores (right) under latent error process $\mu_t = 0, \sigma_t^2=5$}.
    \label{fig:exp2_injected_error_walker.png}
\end{figure}

\begin{figure}[!htb]
    \centering
    \includegraphics[width=1\linewidth]{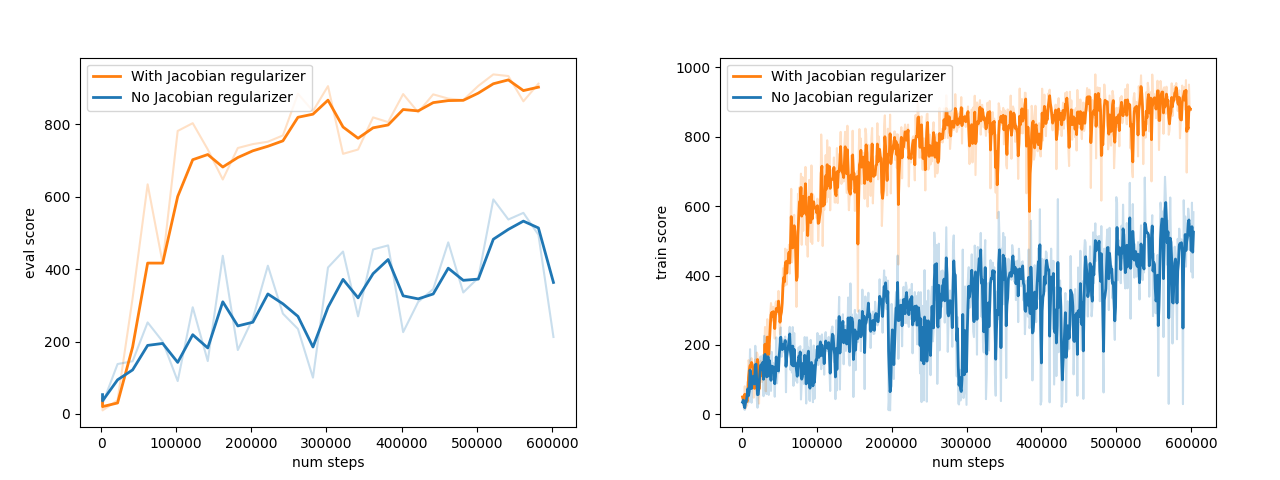}
    \caption{\textit{Walker.} Eval (left) and train scores (right) under latent error process $\mu_t \sim [0, 5], \sigma_t^2 \sim [0,5]$.}
    \label{fig:exp2_injected_error_unif_walker.png}
\end{figure}

Under the Walker task, Figures \ref{fig:exp2_injected_error_walker.png} and \ref{fig:exp2_injected_error_unif_walker.png} show that model with regularization is significantly less sensitive to perturbations in latent state $z_t$ compared to the baseline model without regularization. This empirical observation supports our theoretical findings in Corollary \ref{unstable-imp-thm}, which assert that the impact of latent representation errors on the loss function $\mathcal{L}$ can be effectively controlled by regulating the model’s Jacobian norm.

\newpage
\subsubsection{Quadruped Task}
\begin{figure}[!htb]
    \centering
    \includegraphics[width=1\linewidth]{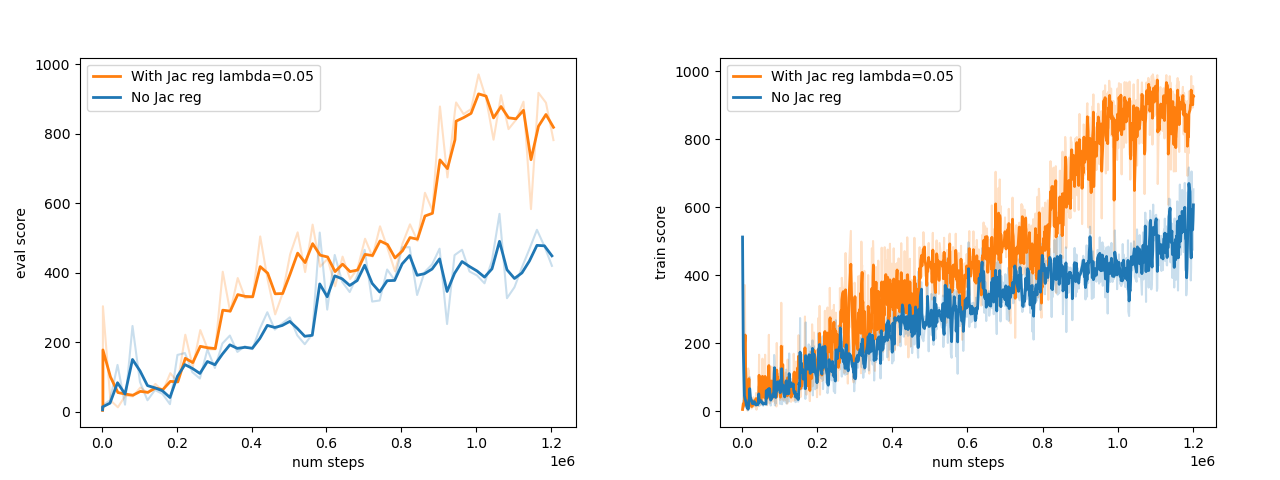}
    \caption{\textit{Quad.} Eval (left) and train scores (right) under latent error process $\mu_t = 0, \sigma_t^2=0.1$.}
    \label{exp2_injected_error_quad.png}
\end{figure}

\begin{figure}[!htb]
    \centering
    \includegraphics[width=1\linewidth]{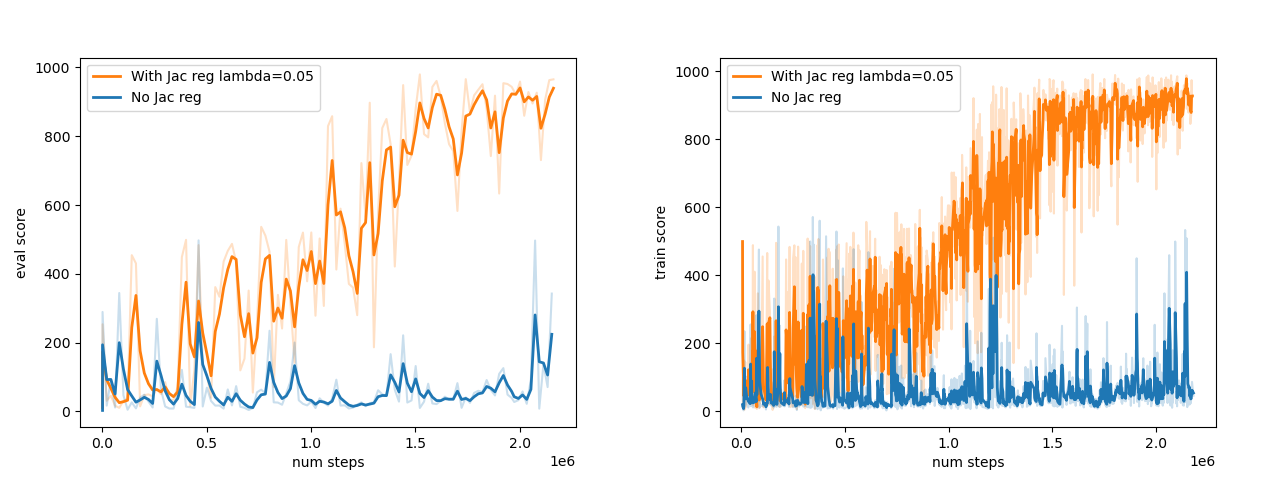}
    \caption{\textit{Quad.} Eval (left) and train scores (right) under latent error process $\mu_t \sim [0, 5], \sigma_t^2 \sim [0,5]$.}
    \label{exp2_injected_error_unif_quad.png}
\end{figure}

Under the Quadruped task,we initially examined a smaller latent error process ($\mu_t = 0, \sigma_t^2=0.1$) and observed that the model with Jacobian regularization converged significantly faster, even though the adversarial effects on the model without regularization were less severe (Figure \ref{exp2_injected_error_quad.png}). When considering the more challenging latent error process ($\mu_t \sim [0, 5], \sigma_t^2 \sim [0, 5]$), we noted that the regularized model remained significantly less sensitive to perturbations in latent state $z_t$, whereas the baseline model struggled to learn (Figure \ref{exp2_injected_error_unif_quad.png}).  These empirical observations reinforce our theoretical findings in Corollary \ref{unstable-imp-thm}, demonstrating that regulating the model’s Jacobian norm effectively controls the impact of latent representation errors.
\newpage

\subsection{Comparison of Jacobian Regularization and Augmentation Methods with Known Perturbation Types
\label{appendix-aug}}

In  cases where additional knowledge about perturbation is available, such as when the perturbation type is known a priori (which could be unrealistic), one could consider using augmentation methods by training with perturbed observations to improve robustness.   
We considered training with observation images augmented with (1) randomly-masked Gaussian noises $\mathcal{N}(0.15, 0.1)$ and (2) rotations $10^\circ$.

\begin{table}[!htb]
{\footnotesize
\centering
\begin{tabular}{c|c|cc|cc|cc}
\toprule
 &   &\multicolumn{2}{c}{full, $\mathcal{N}(0.5, \,\sigma_1^2)$}  & \multicolumn{2}{c}{rotation, $+\alpha^{\circ}$} & \multicolumn{2}{c}{mask $\beta\%$, $\mathcal{N}(0.5, \,0.15)$} \\
\hline
 & clean & $\sigma_1^2 = 0.35$  & $\sigma_1^2 = 0.5$   &$\alpha=20$& $\alpha=30$& $\beta = 50$ & $\beta =75$\\
\hline
Jac Reg & \textbf{967.12}         &    \textbf{742.32}&       \textbf{618.98}&    423.81&      226.04&     725.81& \textbf{  685.49 }\\
Aug w. $\mathcal{N}(0.15, 0.1)$  & 847.19 &   182.33   &  127.72 &       286.63 &    213.93 &               \textbf{767.92}& 187.66   \\
Aug w. rotation $10^\circ$  & 860 &   286.26   &  184.84 &       \textbf{695.34} &   \textbf{ 424.88} &               347.66& 256.84   \\
Baseline  & 966.53 &      615.79&   333.47&       391.65&    197.53&               583.41&    446.74\\
\bottomrule
\end{tabular}
\caption{Evaluation on unseen states by various perturbation (Clean means without perturbation). $\lambda=0.01$.}
\label{table:expr1-table-pert-aug}
\vspace{-14 pt}
}
\end{table}

\begin{table}[!htb]
{\footnotesize
\centering
\begin{tabular}{c|c|c|c|c}
\toprule
 &   g = 9.8 &g = 6  & g = 4 & g = 2 \\
\hline
Jac Reg & \textbf{967.12}         &    \textbf{906.42}&       \textbf{755.18}&    \textbf{ 679.24}\\
Aug w. $\mathcal{N}(0.15, 0.1)$  & 847.19 &  771.34    &624.4  &        	428.45 \\
Aug w. rotation $10^\circ$  & 860 &  582.22    &486.84  &        356.9 \\
Baseline & 966.53 &      750.36 &   662.86&       381.14\\

\bottomrule
\end{tabular}
\caption{Evaluation on unseen dynamics by various gravity constants ($g = 9.8$ is default). $\lambda=0.01$.}
\label{table:expr1-table-grav}
\vspace{-14 pt}
}
\end{table}

As shown in Table \ref{table:expr1-table-pert-aug} and \ref{table:expr1-table-grav},  the experimental results indicate that models trained with Jacobian regularization outperform those using augmentation methods when faced with perturbations different from those used during augmentation. While state augmentation is effective when the inference perturbations match those used in training, it struggles to generalize to unseen perturbations. In contrast, Jacobian regularization is less dependent on the diversity and relevance of augmented data samples, as it directly targets the learning dynamics of the world model. This makes it more broadly applicable and reduces the likelihood of overfitting, avoiding the risk of the model becoming overly specialized to specific perturbation patterns, which is a common challenge with data augmentation.


\newpage
\subsection{Visualizations of reconstructed state trajectory under  exogenous zero-drift and non-zero drift latent representation error.\label{app-vis}}

In this section, we present visualizations of reconstructed state trajectory samples, included in the revision to illustrate the error propagation of exogenous zero-drift and non-zero drift error signals in latent states, both with and without Jacobian regularization.

\begin{figure}[!htb]
    \centering
    \includegraphics[width=\linewidth]{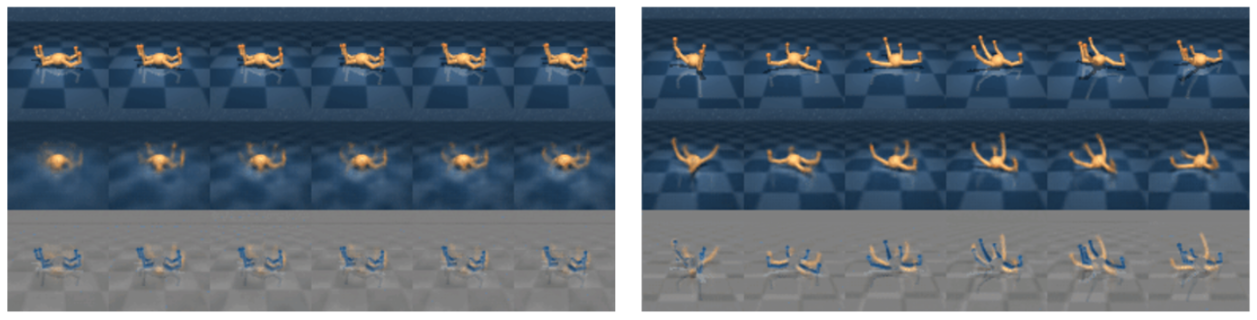}
    \caption{\textit{Quad}. Open-loop reconstructed trajectories under zero-drift latent representation error ($\mu_t = 0, \sigma^2_t = 5$) with \textit{right} and without \textit{left} Jacobian regularization.}
    \label{fig:zero-drift-ol}
\end{figure}

\begin{figure}[!htb]
    \centering
    \includegraphics[width=0.75\linewidth]{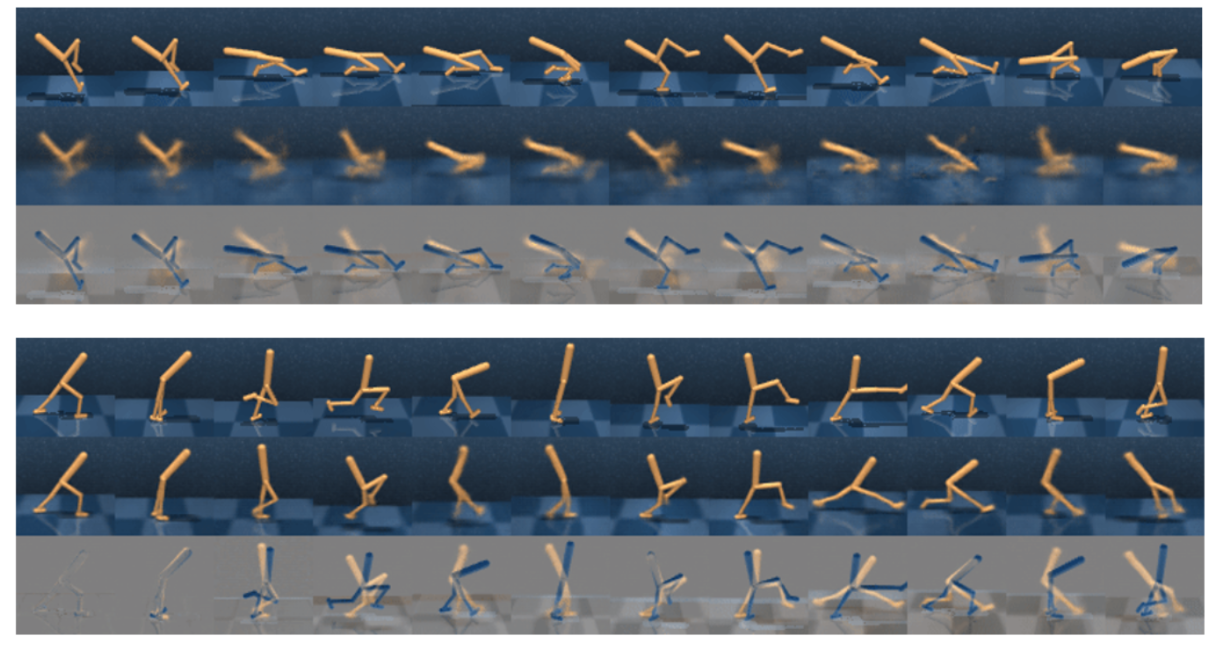}
    \caption{\textit{Walker.} Open-loop reconstructed trajectories under non-zero drift latent representation error 
($\mu_t \sim [0, 5], \sigma^2_t \sim [0, 5]$) with \textit{lower} and without \textit{upper} Jacobian regularization.}
    \label{fig:non-zero-drift-ol}
\end{figure}

As depicted in Figures \ref{fig:zero-drift-ol} and \ref{fig:non-zero-drift-ol}, the reconstructed states for the baseline model without Jacobian regularization appear blurry and less structured, indicating that the model has not effectively captured the underlying dynamics of the environment. In contrast, the reconstructed states for the model with Jacobian regularization are sharper and more accurately reflect the true dynamics of the environment. The visual comparison highlights the robustness brought by Jacobian regularization against latent noises.
\newpage
\subsection{Additional results on faster convergence on tasks with extended horizon.}
\begin{figure}[!htb]
    \centering
    \includegraphics[width=1\linewidth]{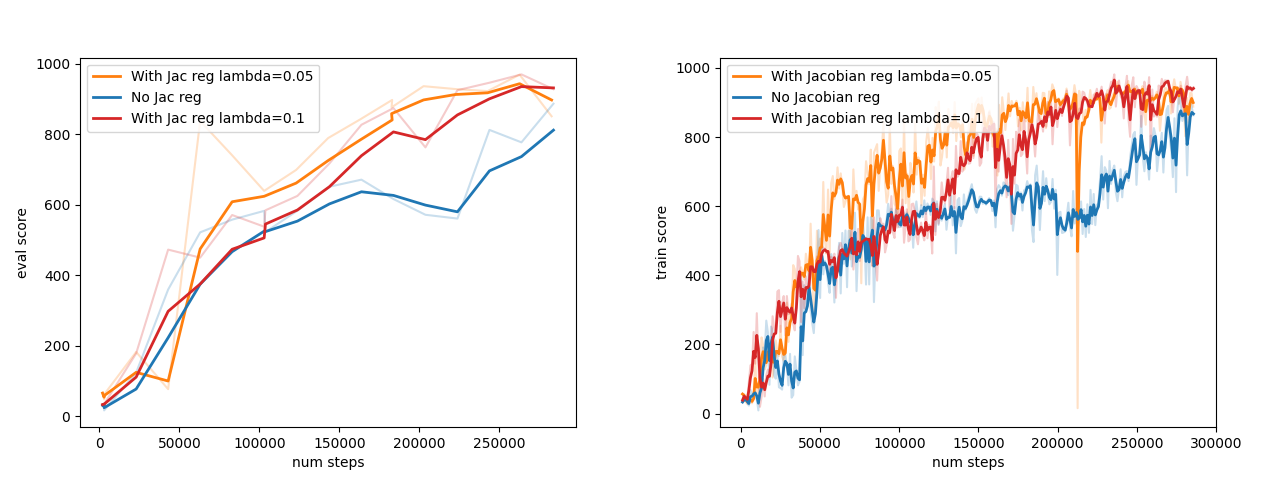}
    \caption{\textit{Extended horizon Walker task.} Eval (left) and train scores (right).}
    \label{exp3_extended_horizon.png}

\caption{\textit{Extended horizon Walker task.} Eval (left) and train scores (right).}
\end{figure}

In this experiment, we evaluate the efficacy of Jacobian regularization in extended horizon tasks, specifically by increasing the horizon length in MuJoCo Walker from 50 to 100 steps. We tested two regularization weights $\lambda=0.1$ and $\lambda=0.05$. Figure \ref{exp3_extended_horizon.png} demonstrates that models with regularization converge faster, with $\lambda=0.05$ achieving convergence approximately 100,000 steps ahead of the model without Jacobian regularization. This supports the findings in Theorem \ref{acc-thm}, indicating that regularizing the Jacobian norm can reduce error propagation, especially over longer time horizons.

\subsection{Additional Details on Batch Size and Robustness (Table 1).}

The batch-size versus robustness experiment (Table 1) uses the same perturbation methods as those employed in the Jacobian regularization experiment with perturbed states (Table 2 and Appendix D.2).

\end{document}